\documentclass{article}

\usepackage{arxiv}

\usepackage[utf8]{inputenc}
\usepackage[T1]{fontenc}

\usepackage[numbers]{natbib}
\usepackage{hyperref}

\usepackage{graphicx}
\usepackage{booktabs}
\usepackage{multirow}
\usepackage{amsmath}
\usepackage{amssymb}
\usepackage{csquotes}
\usepackage{microtype}

\usepackage[format=plain,
            labelfont={bf,it},
            textfont=it]{caption}

\newcommand{\WeightLinear}{\mathbf{w}}

\newcommand{\Concepts}{\mathbf{C}}
\newcommand{\ConceptsDim}{\Concepts \in \mathbb{R}^{n \times p}}

\newcommand{\RealImage}{\mathbf{x}_r}
\newcommand{\SyntheticImage}{\mathbf{x}_s}

\newcommand{\ImagesProjected}{\mathbf{I}}
\newcommand{\ImagesProjectedDim}{\ImagesProjected \in \mathbb{R}^{m \times p}}
\newcommand{\CosineSim}{\mathbf{S}}

\newcommand{\CosineSimFullDim}{\CosineSim = \ImagesProjected  \Concepts^T \in \mathbb{R}^{m \times n}}

\newcommand{\cnnspot}{\textit{CNNSpot}}
\newcommand{\synthbp}{\textit{SynthBuster+}}
\newcommand{\synthb}{\textit{Synthbuster}}
\newcommand{\synthclic}{\textit{SynthCLIC}}
\newcommand{\raisek}{RAISE-1K}

\newcommand{\vocabulary}{C}
\newcommand{\chatgpt}{\vocabulary_{\text{antonyms}}}
\newcommand{\textspan}{\vocabulary_{\text{textspan}}}

\title{Synthetic Image Detection with CLIP: Understanding and Assessing Predictive Cues}

\author{
  Marco Willi \\
  Institute for Data Science I4DS \\
  University of Applied Sciences FHNW\\
  Brugg-Windisch AG, Switzerland \\
  \texttt{marco.willi@fhnw.ch} \\
   \And
  Melanie Mathys \\
  Institute for Data Science I4DS \\
  University of Applied Sciences FHNW\\
  Brugg-Windisch AG, Switzerland \\
  \texttt{melanie.mathys@fhnw.ch} \\
   \And
  Michael Graber \\
  Institute for Data Science I4DS \\
  University of Applied Sciences FHNW\\
  Brugg-Windisch AG, Switzerland \\
  \texttt{michael.graber@fhnw.ch} \\
}

\begin{document}

\maketitle

\begin{abstract}
Recent generative models produce near-photorealistic images, challenging the trustworthiness of photographs.
Synthetic image detection (SID) methods, however, often struggle to generalize across datasets and generative models.
CLIP, which embeds images and text in a shared semantic space, performs well at SID, but the cues underlying its decisions remain poorly understood. We therefore study CLIP-based SID as an empirical interpretability problem rather than proposing a new detector.
We introduce \synthclic{}, which pairs real photographs with caption-matched, high-quality diffusion-generated counterparts. We evaluate CLIP-based detectors on \synthclic{}, a GAN-heavy benchmark, and a broad external benchmark, and compare them with a low-level forensic CNN, a broad-generator detector, and a text-grounded concept model.
CLIP-based linear detectors reach 0.96 mAP on the GAN-heavy benchmark but 0.92 on \synthclic{}, while cross-family transfer to \cnnspot{} falls to 0.42 mAP. Within-class associations between detector scores and text-derived cue scores show that higher synthetic scores correspond to cleaner, more compositionally controlled, and technically polished images, whereas lower scores correspond to messier capture conditions and provenance cues characteristic of real photographs. These associations are distributed across many overlapping cues, and their profiles differ strongly across training datasets. CLIP-based and forensic detectors therefore fail in different ways and provide complementary evidence, while broad generator coverage appears important for robust SID.
\end{abstract}

\keywords{Synthetic image detection \and CLIP \and Deepfake detection \and Multimedia forensics \and Interpretability \and Diffusion models \and Generalization}

\section{Introduction}\label{sec:intro}

The rapid advancement in the quality of image generative models \cite{rombachHighResolutionImageSynthesis2022, podellSDXLImprovingLatent2023,esserScalingRectifiedFlow2024, imagen-team-googleImagen32024, lipmanFlowMatchingGenerative2023} has made it increasingly difficult to distinguish between real and synthetic photographs \cite{luSeeingNotAlways2023, casuAIMirageImpostor2023, nightingaleAIsynthesizedFacesAre2022}.
This enables nefarious actors to create convincing material to deceive and manipulate their targets, for example in disinformation campaigns or cyber influence operations \cite{chesneyDeepFakesLooming2019, cordeyCyberInfluenceOperations2019, faridCreatingUsingMisusing2022}. Consequently, there is a growing body of research focused on detecting synthetic imagery \cite{marraGANsLeaveArtificial2019, wangCNNgeneratedImagesAre2020, cozzolinoZeroShotDetectionAIGenerated2025, tariangSyntheticImageVerification2024} to mitigate such risks.

Synthetic image detection (SID) methods often exploit model-specific \textit{fingerprints} \cite{marraGANsLeaveArtificial2019}---subtle traces originating from the generative process---to identify synthetic images. Supervised learning algorithms can be trained to recognize these fingerprints with high accuracy \cite{wangCNNgeneratedImagesAre2020}. Such approaches, however, suffer from two major limitations: (i) their accuracy is often tightly coupled to the specific generative model used for model training \cite{zhangDetectingSimulatingArtifacts2019}, and (ii) they are vulnerable to adversarial attacks that remove or obscure these low-level cues~\cite{carliniEvadingDeepfakeImageDetectors2020}. Consequently, SID methods require constant re-training and have unknown effectiveness in open-set conditions where test-time generative models might differ from those seen during training \cite{epsteinOnlineDetectionAIGenerated2023} (referred to as generalization).

A recent line of work investigates the utility of foundational vision models---trained with self-supervised objectives on internet-scale datasets---for synthetic image detection. Particularly, the CLIP model \cite{radfordLearningTransferableVisual2021} has been a focus of intensive research~\cite{ojhaUniversalFakeImage2023, cioniAreCLIPFeatures2024, huangASAPInterpretableAnalysis2024a, khanCLIPpingDeceptionAdapting2024, gaintsevaImprovingInterpretabilityRobustness2024, koutlisLeveragingRepresentationsIntermediate2024, cozzolinoRaisingBarAIgenerated2024a, derosaExploringAdversarialRobustness2024}.
CLIP is trained on large-scale datasets of image-text pairs with the objective of maximizing similarity between matching pairs in a joint text-vision embedding space. The semantic richness of CLIP’s visual encoder, combined with its robustness to distribution shifts \cite{radfordLearningTransferableVisual2021}, makes it a promising candidate for detecting synthetic images.

Prior results suggest that CLIP-based detectors do not rely on low-level forensic fingerprints \cite{cozzolinoRaisingBarAIgenerated2024a}---which is plausible given they are trained to match semantic image descriptions.

Naturally, this raises a fundamental question: \textit{Which photographic, semantic, and forensic information represented by CLIP supports SID decisions?} Furthermore, \textit{how effective are CLIP-based methods considering generative models continue to improve in fidelity and diversity?}

\begin{figure}[htbp]
    \centering
    \includegraphics[trim=4 68 4 36,clip,width=\linewidth]{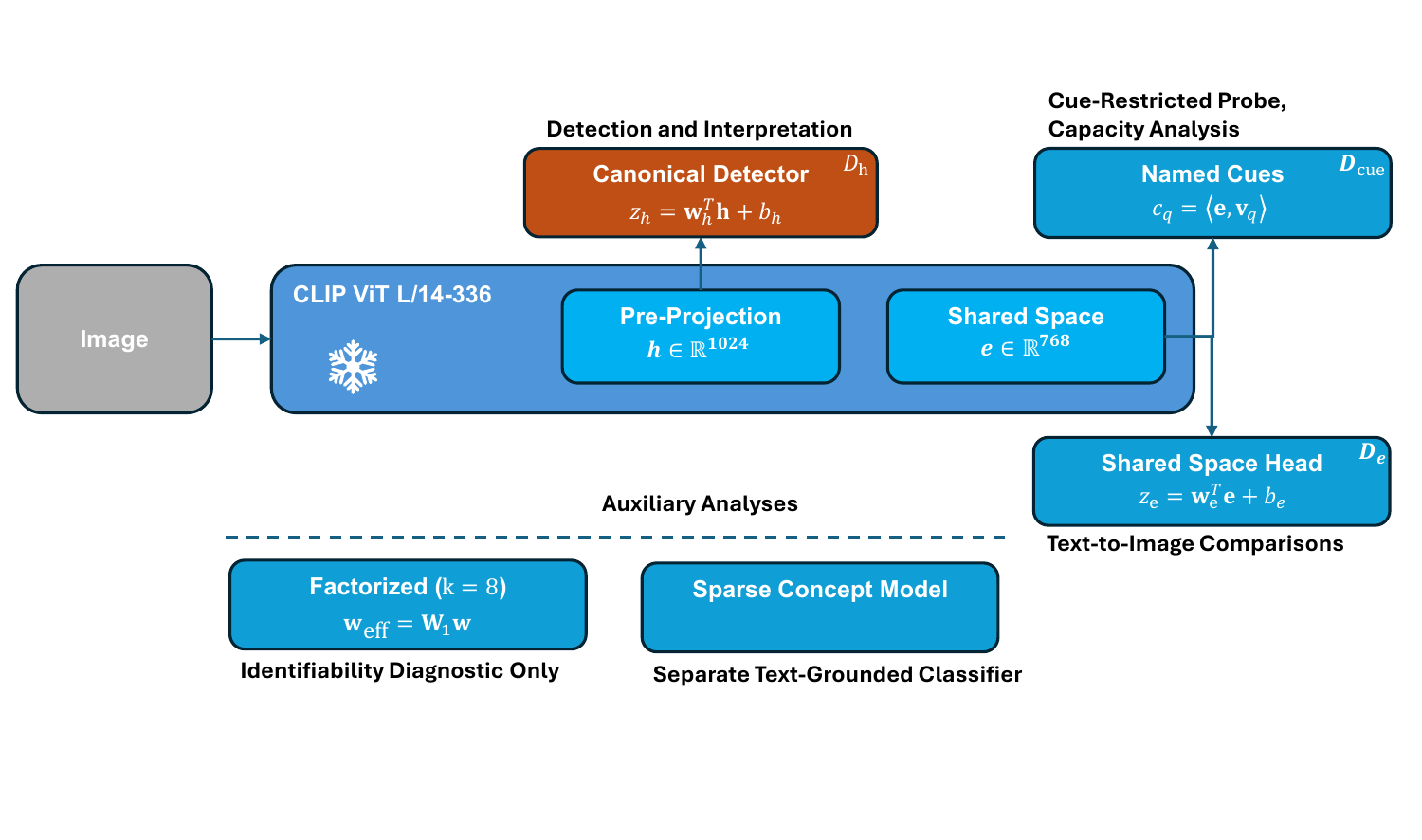}
    \caption{Architecture overview. CLIP (frozen weights) computes the $1024$-d pre-projection visual representation $\mathbf{h}$ for the canonical detector $D_h$. Projection yields the $768$-d shared image--text representation $\mathbf{e}$ for text-coordinate analyses through the shared-space analysis head $D_e$. We evaluate a cue-restricted probe $D_{\mathrm{cue}}$ on 168 cue scores. The factorized $k{=}8$ head is a diagnostic attempt to isolate $k$ interpretable projections, and the sparse concept model is a separate text-grounded classifier.}
    \label{fig:intro}
\end{figure}

Our goal is to identify which cue families in CLIP space support SID decisions and whether these cues are stable under changes in model initialization and dataset. We treat SID performance as a prerequisite for interpretation, not as the main objective. Specifically, we address the following questions:

\begin{itemize}
    \item [Q1:] How do CLIP-based SID detectors behave across paired diffusion datasets, GAN-heavy benchmarks, and cross-family evaluation settings?
    \item [Q2:] Which explanation units for CLIP-based SID are stable and interpretable: individual axes of a factorized linear head, the effective linear decision direction, or sparse text-grounded concepts?
    \item [Q3:] Which photographic, perceptual, or forensic cue families in CLIP space distinguish real from synthetic images, and how do these cues change across datasets and generator families?
\end{itemize}

We seek to answer these questions by systematically analyzing the behavior and capabilities of CLIP-based synthetic image detectors. We leverage CLIP’s cross-modal nature to probe its visual representations with textual descriptions (cues), offering an interpretable window into what drives its predictions. Additionally, we evaluate CLIP-based detectors across datasets of varying generative quality to understand their performance boundaries and failure modes.

Our contributions are fourfold:

\begin{enumerate}

    \item %
    We introduce \synthclic{}, a paired dataset of real photographs from the \textit{CLIC} dataset and high-quality synthetic counterparts from recent diffusion models, designed to support content-controlled analysis of CLIP-based SID cues.

    \item %
    We evaluate the performance of CLIP-based SID across paired diffusion datasets and a GAN-heavy benchmark. We compare with low-level forensic detectors and a broad-generator out-of-the-box detector for reference.

    \item %
    We show that explanations based on individual axes of an over-parameterized k-dimensional linear head are not constant across random initializations. We therefore use single linear decision directions and photographic cues as the primary interpretation units.

    \item %
    We analyze CLIP-based SID decisions with CLIP-IQA-style perceptual attributes, paired content-controlled cue shifts, and score-space alignment measured directly against the evaluated checkpoints, and separately evaluate text-grounded concept models as complementary vocabulary-conditioned classifiers rather than as explanations of the linear detector. Figure~\ref{fig:intro} summarizes this detector and analysis hierarchy.
    
\end{enumerate}

\section{Previous Work}\label{sec:related}

\subsection{Generative Modeling}
\label{sec:prev_work:gen}

Generative models learn the distribution of a dataset and enable sampling of new data points. Often, such models allow for conditional generation, for example, to condition on class labels or textual descriptions. Recent advances in generative modeling have enabled near photo-realistic creation of images. There are different model types, such as generative adversarial networks (GANs) \cite{Goodfellow2014}, variational autoencoders (VAEs) \cite{kingmaAutoEncodingVariationalBayes2013}, normalizing flows \cite{dinhDensityEstimationUsing2017}, autoregressive models \cite{yuScalingAutoregressiveModels2022, oordPixelRecurrentNeural2016}, and diffusion models (DMs) \cite{hoDenoisingDiffusionProbabilistic2020}. In particular, DMs have demonstrated impressive generative quality, such as the stable diffusion family of models \cite{rombachHighResolutionImageSynthesis2022, podellSDXLImprovingLatent2023, esserScalingRectifiedFlow2024}. Furthermore, there are commercial providers which offer access to proprietary (and sometimes open-source) models, such as Midjourney \cite{Midjourney}, Google's Imagen3 \cite{imagen-team-googleImagen32024}, Black Forest's Flux models \cite{AnnouncingBlackForest2024}, Adobe's Firefly models \cite{AdobeFirefly}, or OpenAI's GPT-Image \cite{Introducing4oImage2025}. The most recent advances include multi-modal large language models which can be used to generate images and text consistently and in conditional sequences \cite{yinSurveyMultimodalLarge2024, zhouTransfusionPredictNext2024}, including image captioning.

\subsection{Synthetic Image Detection (SID)}
\label{sec:prev_work:sid}

Generative models leave invisible traces---also referred to as \textit{artificial fingerprints}---in generated images, which are tied to the generative process \cite{marraGANsLeaveArtificial2019}. These fingerprints can be visualized by calculating the average frequency spectra of images from generative models \cite{zhangDetectingSimulatingArtifacts2019, marraGANsLeaveArtificial2019}. This has been demonstrated for different models and architecture types, including diffusion models \cite{corviIntriguingPropertiesSynthetic2023}. A common approach to SID is to model it as a binary classification task using a dataset of real and synthetic images. Several works \cite{wangCNNgeneratedImagesAre2020, corviDetectionSyntheticImages2023, rickerDetectionDiffusionModel2024, shaDEFAKEDetectionAttribution2023, wangDIREDiffusionGeneratedImage2023, wongLocalStatisticsGenerative2023, wuGeneralizableSyntheticImage2023} demonstrated that this approach can achieve high accuracy. Models trained for SID tend to generalize poorly to synthetic images from models not seen during model training \cite{cozzolinoForensicTransferWeaklysupervisedDomain2019}. Generalization improves with aggressive data augmentation during model training, such as post-processing operations like image compression \cite{wangCNNgeneratedImagesAre2020}. However, models do not generalize across substantial architectural changes, e.g., from GANs to Diffusion models \cite{epsteinOnlineDetectionAIGenerated2023}.

A complementary strategy is to broaden training coverage rather than change the feature space. Recent face-forgery research has addressed distribution shift through multi-domain learning and more realistic data generation. GM-DF \cite{lai_gm-df_2025} jointly models heterogeneous forgery datasets and uses CLIP to represent features common to several domains, while Agent4FaceForgery \cite{lai_agent4faceforgery_2026} generates ecologically diverse face-forgery data through multi-agent simulation. Although these works address face manipulation rather than fully synthetic images, they reinforce the importance of heterogeneous training data. More directly related to our setting, CommunityForensics \cite{park_community_2025} trains on images from thousands of generators and demonstrates strong cross-generator transfer. We therefore include CommunityForensics-384 as a broad-exposure reference detector and evaluate all detectors on its evaluation set (CF-Eval) alongside our own test splits.

Additionally, SID models remain vulnerable to adversarial attacks, as shown in both white-box and black-box scenarios \cite{carliniEvadingDeepfakeImageDetectors2020, zhouStealthDiffusionEvadingDiffusion2024}. Furthermore, optimal detection thresholds vary for accurate detection of synthetic images from different generative models, making it necessary to calibrate model predictions when doing inference on synthetic images from novel sources.

\subsection{CLIP-based Synthetic Image Detection}

A growing body of work demonstrates that CLIP's image representations are remarkably effective for SID~\cite{ojhaUniversalFakeImage2023, cioniAreCLIPFeatures2024, huangASAPInterpretableAnalysis2024a, khanCLIPpingDeceptionAdapting2024, gaintsevaImprovingInterpretabilityRobustness2024, koutlisLeveragingRepresentationsIntermediate2024, cozzolinoRaisingBarAIgenerated2024a, derosaExploringAdversarialRobustness2024}. A common paradigm is to extract features using CLIP's frozen image encoder and train a lightweight classifier—typically logistic regression or a linear SVM—on top of those features. Across studies, CLIP-based representations consistently outperform features from supervised models in both accuracy and generalization.

A central insight from this line of work is the effectiveness of contrastive pre-training combined with the vision transformer (ViT) architecture \cite{dosovitskiyImageWorth16x162020}, as opposed to CNN-based backbones. CLIP-ViT embeddings not only outperform supervised ImageNet features for SID but also naturally separate real from synthetic images in 2D t-SNE projections, even without fine-tuning \cite{ojhaUniversalFakeImage2023}. These findings demonstrate the inherent discriminative capacity of CLIP representations for SID.

Another important thread investigates the layer-wise utility of CLIP representations. While most studies use (near) final image embeddings, several works show that intermediate layers contain richer low-level cues crucial for detecting artifacts and visual fingerprints. Fusing intermediate and final-layer features with a learnable module yields substantial gains \cite{koutlisLeveragingRepresentationsIntermediate2024}; intermediate features also outperform early or late layers in open-set and out-of-distribution scenarios, particularly with limited training data \cite{cioniAreCLIPFeatures2024}.

A third theme is robustness and data efficiency. CLIP-based detectors generalize to unseen generative models and remain stable under perturbations such as compression and resizing; they achieve strong performance with as few as 10 training examples, with diminishing returns beyond roughly 10,000 image pairs. Mitigating semantic biases via paired training---generating a synthetic counterpart from a real image’s caption---further improves reliability \cite{cozzolinoRaisingBarAIgenerated2024a}.

Finally, joint adaptation of CLIP’s visual and textual components has been explored. Comparisons of linear probing, fine-tuning, prompt tuning, and adapters identify prompt tuning as most effective in many settings; leveraging the text encoder through systematic input adaptation can further enhance SID, though results vary across datasets \cite{khanCLIPpingDeceptionAdapting2024}.

\subsection{Why are CLIP-based Detectors Effective?}

CLIP models are pre-trained on massive datasets of image-text pairs—400 million in the original release~\cite{radfordLearningTransferableVisual2021}, and over 2 billion in recent variants~\cite{chertiReproducibleScalingLaws2023, schuhmannLAION5BOpenLargescale2022}. This scale and diversity enable CLIP to learn rich, generalizable visual features by aligning images with textual descriptions. Models trained from scratch in a supervised setting may focus on the (easy) task of detecting generative fingerprints and thus ignore potential features related to real photographs \cite{ojhaUniversalFakeImage2023}. This behavior could lead to poor generalization: in the absence of model-specific fingerprints everything gets classified as real. Models not explicitly trained for SID might provide a richer, and better suited feature-space \cite{ojhaUniversalFakeImage2023}.

Previous work \cite{cozzolinoRaisingBarAIgenerated2024a} showed that CLIP-based detectors retain high performance under image manipulations such as resizing and compression, which often degrade the subtle traces exploited by conventional detectors. They found that supervised models (CNNs trained from scratch) fail when (i) synthetic images are resized to suppress high-frequency artifacts, or (ii) real images are modified using an autoencoder to embed artificial fingerprints. In contrast, CLIP-based detectors remain effective, suggesting their decisions are based on features beyond low-level image statistics.

A study \cite{derosaExploringAdversarialRobustness2024} examined the adversarial robustness of CLIP- and CNN-based SID models. Their Fourier-domain analysis revealed that adversarial perturbations targeting CLIP tend to affect a wider frequency range, particularly lower components, than those targeting CNNs. Consistently, they observed that low-pass filtering disproportionately harms CNN-based detectors. Overall, this is indicative of larger image structures being involved in CLIP-based SID approaches. Although CLIP is not inherently more robust to adversarial attacks, adversarial examples rarely transfer between CLIP and CNN architectures, underscoring fundamental differences in the features they exploit.

\subsection{Interpreting CLIP-based Models}
\label{sec:prev_work:interpreting_clip}

CLIP embeds images and text in a shared space, enabling zero-shot classification and geometric analyses (e.g., linear directions aligned with textual prompts) \cite{radfordLearningTransferableVisual2021}.
For SID, \cite{gaintsevaImprovingInterpretabilityRobustness2024} aligns linear-probe weights with text embeddings, prunes redundant features, and inspects attention heads. This reveals blur-related cues and dataset artifacts as being relevant features for CLIP-based SID; tokens like \enquote{blurry}/\enquote{blur} correlate with specific image generators, while \enquote{deeplearning}/\enquote{generative} associate with GANs—likely reflecting captions in CLIP’s pretraining data. Experiments span GAN-based datasets and paired images from several text-conditional diffusion models.
Complementary research, unrelated to SID, reports head-level specialization in ViT-based CLIP and introduces \textit{TextSpan}, which constructs a semantic basis for each attention head by matching activations to text embeddings \cite{gandelsmanInterpretingCLIPsImage2024}. A related but distinct line of work, concept discovery models \cite{panousisSparseLinearConcept2023}, links a set of human-interpretable concepts to images by introducing per-example binary concept-selection latents (with a Bernoulli prior) and learning them end-to-end by maximizing an ELBO.

\subsection{Relation to prior CLIP-based SID}

Prior work has demonstrated the effectiveness of CLIP features for SID, typically by training linear classifiers or shallow heads on top of CLIP’s final image embedding and evaluating on specific benchmarks and generator families. In contrast, we (i) evaluate CLIP-based SID across three datasets including a new paired dataset (\synthclic{}) with high-quality diffusion-based images derived from CLIC,

(ii) we analyze stable linear decision directions and the associated text-grounded cue profiles, and explicitly test the stability of different explanation units, showing that the effective ($k{=}1$) decision direction is more reliable than individual axes of an over-parameterized factorized head,
and (iii) adapt sparse linear concept discovery models to CLIP-based SID using photography-oriented vocabularies. This allows us not only to quantify detection performance but also to investigate which visual and photographic attributes are associated with CLIP-based decisions about whether an image is real or synthetic.

\section{Method}\label{sec:method}

\subsection{Image Data}
\label{sec:method:image_data}

We use three different datasets for our experiments. The first, \cnnspot{}, was introduced by \cite{wangCNNgeneratedImagesAre2020} and is used in many subsequent works for SID, including CLIP-based approaches \cite{ojhaUniversalFakeImage2023}. The dataset consists of a training set based on synthetic images generated solely by the ProGAN method, which was trained to generate 20 different classes from the LSUN \cite{yuLSUNConstructionLargescale2016} dataset. The test set of \cnnspot{} is more diverse and includes images from 11 different generative models, mostly based on GANs. For efficiency reasons and because previous work (e.g. \cite{cozzolinoRaisingBarAIgenerated2024a}) demonstrated low sensitivity to the number of training images, we only use 1k training images, instead of the full dataset with 720k images. Example images can be found in Appendix~\ref{appendix:datasets_example}.

\begin{figure}[htbp]
    \centering
    \includegraphics[width=1.0\linewidth]{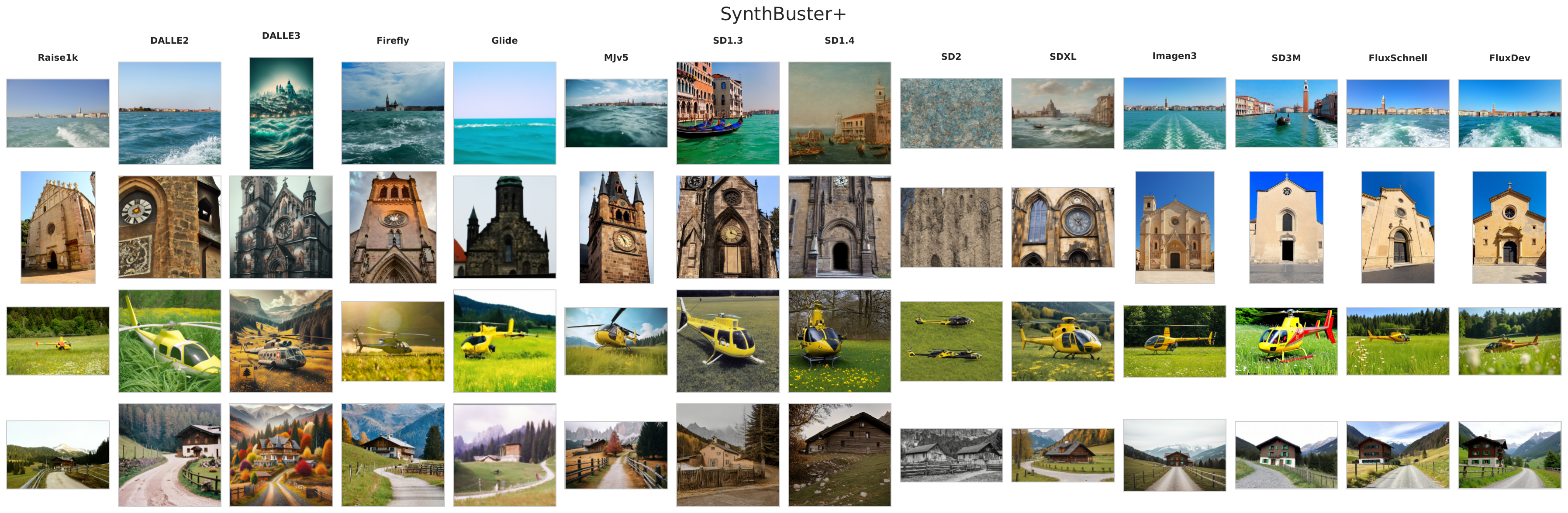}
    \caption{Examples from the \synthbp{} dataset. Different paired images are shown in each row. Each column depicts a different image source, starting with real photographs from the \raisek{} dataset \cite{dang-nguyenRAISERawImages2015}, followed by synthetic images from the \synthb{} dataset \cite{bammeySynthbusterDetectionDiffusion2023} and images added by us: Imagen 3 \cite{imagen-team-googleImagen32024}, FluxDev and FluxSchnell \cite{AnnouncingBlackForest2024}, and Stable Diffusion 3 Medium \cite{esserScalingRectifiedFlow2024}.}
    \label{fig:data:synthbuster_plus}
\end{figure}

Additionally, we use the \textit{SynthBuster} dataset \cite{bammeySynthbusterDetectionDiffusion2023} which is based on real photographs (\raisek{}) \cite{dang-nguyenRAISERawImages2015} and corresponding, paired synthetic images which were generated with text-conditional generative models to match the content of the real photographs as closely as possible. The dataset includes images from nine different generative latent diffusion models. We augment the dataset with synthetic images from more recent models to reflect the current generative quality and name the dataset \synthbp{}. Qualitative examples illustrating the dataset are provided in Figure~\ref{fig:data:synthbuster_plus}.

\begin{figure}[htbp]
    \centering
    \includegraphics[width=0.5\linewidth]{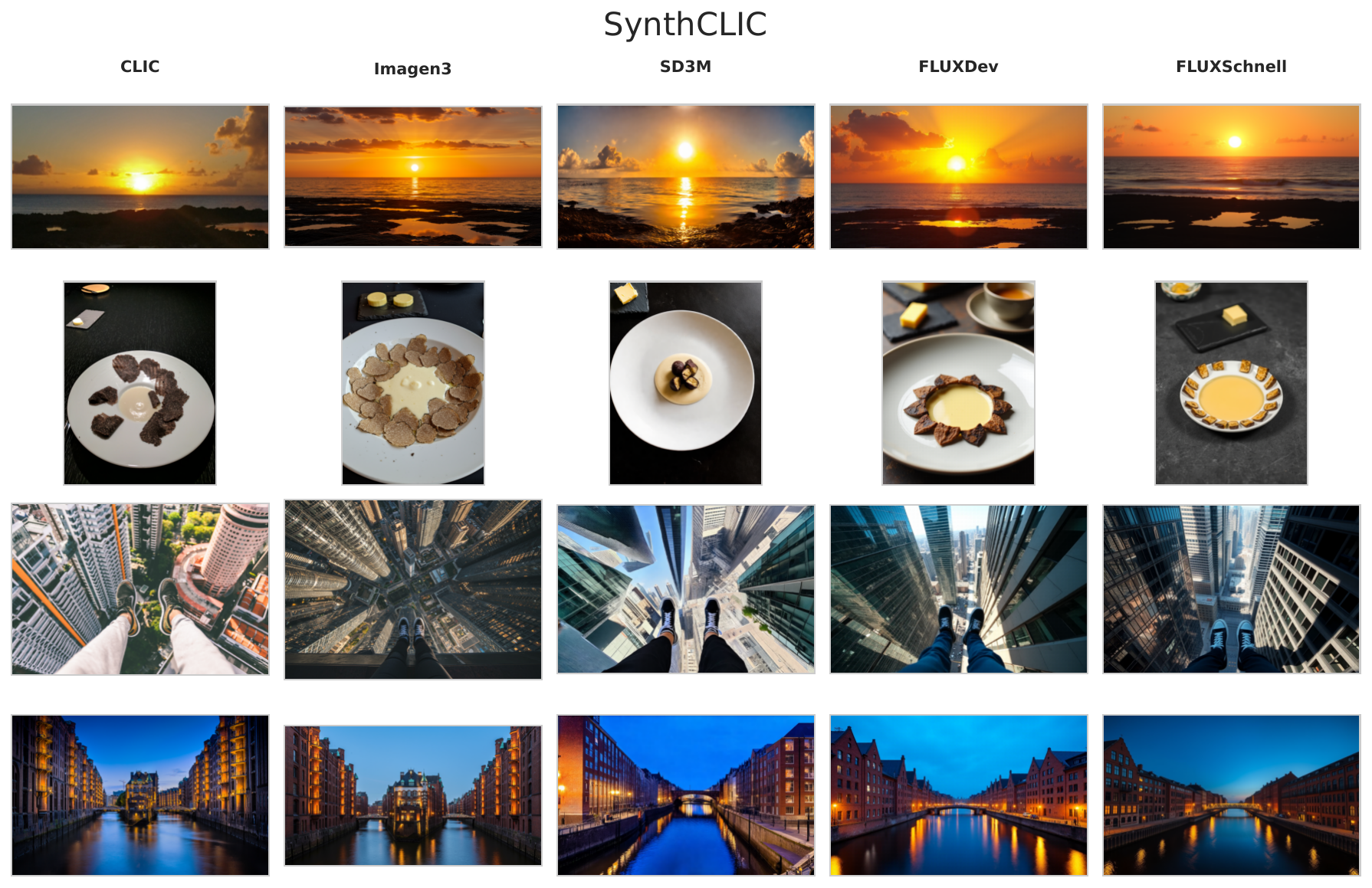}
    \caption{Examples from the \synthclic{} dataset. Different paired images are shown in each row. Each column depicts a different image source, starting with real photographs from the CLIC dataset, followed by synthetic images generated with Imagen 3 \cite{imagen-team-googleImagen32024}, FluxDev and FluxSchnell \cite{AnnouncingBlackForest2024}, and Stable Diffusion 3 Medium \cite{esserScalingRectifiedFlow2024}.}
    \label{fig:data:synthclic}
\end{figure}

We observe that the real photographs used in \textit{SynthBuster} are not quite recent and lack diversity, with a strong bias toward outdoor and street scenes. Moreover, many real images appear overly bright and of relatively low photographic quality. To mitigate the risk that detectors exploit these biases, we construct an additional dataset, which we call \synthclic{}. As a basis, we use real photographs from the 6th Challenge on Learned Image Compression (CLIC)\footnote{\href{http://www.compression.cc/}{http://www.compression.cc/}}. These images include both professionally captured photographs and images taken with mobile phones. We then use the same generative models as those used to extend the \textit{SynthBuster} dataset to generate corresponding synthetic images. Figure \ref{fig:data:synthclic} shows some example images.

We create images using Imagen 3~\cite{imagen-team-googleImagen32024}, FluxDev and FluxSchnell~\cite{AnnouncingBlackForest2024}, and Stable Diffusion 3 Medium~\cite{esserScalingRectifiedFlow2024} (see Appendix \ref{appendix:datasets_genmodels} for exact model versions used). We aim at preserving the aspect ratio of each corresponding real image. We condition the generation process on descriptions derived from Gemini \cite{comaniciGemini25Pushing2025}, a commercial visual large language model. Table \ref{tab:data:sizes} provides an overview of the datasets, including the split sizes used for model training, validation (a subset of the training set, used for early stopping), and testing. In paired datasets any real image and its synthetic variants are assigned to the same split.

\begin{table}[!ht]
\centering
\scriptsize
\begin{tabular}{lrrrrrr}
\toprule
 & \multicolumn{2}{c}{Train} & \multicolumn{2}{c}{Validation} & \multicolumn{2}{c}{Test} \\
 \midrule
 & Real & Synthetic & Real & Synthetic & Real & Synthetic \\
\midrule
CNNSpot & 1000 & 1000 & 1000 & 1000 & 54970 & 52980 \\
SynthBuster+ & 640 & 8319 & 160 & 2080 & 200 & 2600 \\
SynthCLIC & 1633 & 6532 & 102 & 408 & 428 & 1712 \\
\bottomrule
\end{tabular}
\caption{Datasets and the number of real and synthetic images, including split sizes, used in subsequent experiments. For models trained on all datasets (\textit{combined}) the size is the sum over the individual datasets.}
\label{tab:data:sizes}
\end{table}

To characterize and compare the synthetic images used in our experiments, and in particular to characterize potential perceptual differences and dataset biases, we employed a method from Image Quality Assessment (IQA). Such methods aim to quantify the \textit{look} and \textit{feel} of images from a human perspective. We used CLIP-IQA \cite{wangExploringCLIPAssessing2022}, a CLIP-based approach, to score several photographic attributes. The resulting score distributions show limited overlap for some attributes, such as \textit{quality}, \textit{beautiful}, \textit{noisiness}, and \textit{sharpness} (see Appendix~\ref{appendix:clip_iqa}), suggesting that CLIP-based detectors can, to some extent, exploit these cues to distinguish real from synthetic images. At the same time, visual inspection of paired images reveals strong semantic similarity between real and synthetic versions (Figures~\ref{fig:data:synthbuster_plus} and \ref{fig:data:synthclic}).

\subsection{Evaluation Settings and Metrics}

We consider three evaluation settings. (i) In-dataset (closed-set) detection: train and test on the same dataset, which measures performance when the detector is matched to both the real-image distribution and the set of generators used in that dataset. (ii) Cross-dataset generalization with overlapping generators: train on one diffusion-based dataset (\synthbp{} or \synthclic{}) and evaluate on the other. In this case, many underlying diffusion models overlap, but the real-image distributions, prompts, and generation pipelines differ, so the task mainly probes robustness to content and distribution shift rather than to completely unseen generators. (iii) Cross-family generalization: train on the GAN-heavy \cnnspot{} benchmark and evaluate on the diffusion-focused datasets, and vice versa, which combines both a change in real-image distribution and a shift in generator family.

For detector performance, we compute AP per generator using its synthetic images and corresponding real-image set and report the mean over generators (mAP), following common SID practice. Interpretation analyses instead use AUROC for matched comparisons among $D_h$, $D_e$, and $D_{\mathrm{cue}}$, and correlations or paired effect sizes for cue-association and paired-shift analyses. Confidence intervals for the paired comparisons $D_{\mathrm{cue}}$ versus $D_e$ and $D_h$ versus $D_e$ use a cluster bootstrap over source photographs, retaining all associated synthetic variants within each resampled cluster.

\subsection{CLIP Representations}
\label{sec:method:clip_representations}

We use a pre-trained CLIP model, the ViT-L/14-336 variant\footnote{\href{https://huggingface.co/openai/clip-vit-large-patch14-336}{https://huggingface.co/openai/clip-vit-large-patch14-336}}, based on the vision transformer architecture \cite{dosovitskiyImageWorth16x162020}. Let $L$ denote the final transformer layer of the CLIP vision encoder.

For an image $\mathbf{x}^{(i)}$, we use its layer-normalized \texttt{[CLS]} representation
\begin{equation}
\mathbf{h}^{(i)}
=
\mathbf{a}^{L,(i)}
\in
\mathbb{R}^{1024}.
\label{eq:clip_hidden_representation}
\end{equation}
CLIP subsequently maps this representation through its learned image projection
\begin{equation}
\mathbf{P}
\in
\mathbb{R}^{768 \times 1024}
\end{equation}
to the normalized shared image--text representation
\begin{equation}
\mathbf{e}^{(i)}
=
\frac{
    \mathbf{P}\mathbf{h}^{(i)}
}{
    \left\lVert \mathbf{P}\mathbf{h}^{(i)} \right\rVert_2
}
\in
\mathbb{R}^{768}.
\label{eq:clip_projected_representation}
\end{equation}

We distinguish the $1024$-dimensional pre-projection visual representation $\mathbf{h}$ from the $768$-dimensional shared image--text representation $\mathbf{e}$. We do so because the CLIP projection is optimized for cross-modal alignment and may discard visual information relevant to synthetic image detection. Accordingly, our canonical detector operates directly on $\mathbf{h}$, while $\mathbf{e}$ is used for analyses that require direct comparison with text embeddings. All image representations are produced by the pre-trained CLIP encoder, which remains frozen throughout.

\subsection{Canonical CLIP Detector}
\label{sec:method:clip_linear_classifier}

The evaluated detector is a linear binary classifier on the pre-projection visual representations,
\begin{equation}
z^{(i)} = \WeightLinear^{\top} \mathbf{h}^{(i)} + b,
\qquad
\hat{p}^{(i)} = \sigma\bigl(z^{(i)}\bigr),
\label{eq:clip_linear_logit}
\end{equation}
with synthetic-class logit $z^{(i)}$ and predicted probability $\hat{p}^{(i)}$. Over the $N$ labeled training samples we minimize
\begin{equation}
\mathcal{L} = \frac{1}{N}\sum_{i=1}^{N} \mathrm{BCE}\bigl(\hat{p}^{(i)},\, y^{(i)}\bigr),
\end{equation}
where $y^{(i)} \in \{0,1\}$ is the label of sample $i$ and $y^{(i)} = 1$ denotes synthetic. The weight vector $\WeightLinear$ defines this detector's decision boundary. This is the canonical CLIP detector used for all primary CLIP detector-performance results in the manuscript.

We use deterministic pre-processing, weight decay, and early stopping for regularization. We do not use random image augmentation in the primary experiments, so that the learned direction corresponds to a fixed CLIP representation of the evaluated image distribution. Some training splits are substantially imbalanced (Table~\ref{tab:data:sizes}); we do not apply class re-weighting or re-sampling, and therefore emphasize threshold-independent metrics such as mAP. See Appendix~\ref{appendix:method:clip_classifier_implementation_details} for implementation details.

We also experimented with a factorized, two-layer linear head whose activation-level penalty encourages orthogonality over the observed CLIP features (Appendix~\ref{appendix:clip:orthogonal_head}, Equation~\ref{eq:activation_orthogonality}). Without a nonlinearity it spans the same class of linear decision boundaries as Equation~\ref{eq:clip_linear_logit}. Refits show that its individual axes depend on initialization and are unstable for interpretation (Section~\ref{sec:results:interpretation_target}), so we treat it as a diagnostic rather than the canonical detector.

The approach is closely related to prior CLIP-based SID work: \cite{ojhaUniversalFakeImage2023} used a functionally equivalent single linear layer, and \cite{cozzolinoRaisingBarAIgenerated2024a} used the same frozen CLIP vision features with an SVM.

\subsection{Text-Grounded Vocabularies}
\label{sec:method:vocabularies}
We use two vocabularies to interpret SID with CLIP: (i) \textit{TextSpan} (3{,}498 descriptions, $C_{\text{textspan}}$) from \cite{gandelsmanInterpretingCLIPsImage2024}; and (ii) a photography-focused antonym vocabulary (168 attributes, $\chatgpt$) consisting of paired attribute descriptors (e.g., \emph{sharp detail} $\leftrightarrow$ \emph{blurry detail}) with corresponding prompt templates. For each attribute, we embed the positive and negative prompts with CLIP, subtract the normalized negative embedding from the normalized positive embedding, and apply $\ell_2$ normalization. Candidate attributes and antonym pairs were generated using ChatGPT with a prompt requesting photographic and perceptual cues (e.g., color, lens effects, technique), while discouraging object semantics. See Table \ref{tab:vocab_antonyms_example} for examples (prompts not shown, only attributes).

We refer to each signed attribute direction $\mathbf{v}_q$ as a \textit{cue}, and to its projection on an image embedding $c_q = \bigl\langle \mathbf{e},\, \mathbf{v}_q \bigr\rangle$ as the corresponding \textit{cue score}.
Writing $\mathbf{t}_q^{\pm}$ for the normalized positive- and negative-prompt embeddings of attribute $q$, this yields the cue direction
\begin{equation}
\mathbf{v}_q = \frac{\mathbf{t}_q^{+} - \mathbf{t}_q^{-}}{\lVert \mathbf{t}_q^{+} - \mathbf{t}_q^{-} \rVert}.
\label{eq:cue_direction}
\end{equation}
Both vocabularies are fixed in advance and not optimized for SID.

\begin{table}[htbp]
\centering
\scriptsize
\begin{tabular}{@{}p{1.5cm}p{2cm}p{2cm}p{1.2cm}@{}}
\toprule
\textbf{Attribute} & \textbf{Positive} & \textbf{Negative} & \textbf{Category} \\
\midrule
saturation & balanced saturation & oversaturation & color \\
vibrance & high vibrance & muted palette & color \\
vignetting & subtle vignetting & heavy vignetting & lens \\
glitch artifacts & no glitch artifacts & glitch artifacts & other \\
\bottomrule
\end{tabular}
\caption{Example vocabulary antonym pairs used for constructing $\chatgpt$. Each pair yields one cue direction via Equation~\ref{eq:cue_direction}.}
\label{tab:vocab_antonyms_example}
\end{table}

\subsection{Detector Interpretation}
\label{sec:method:interpreting_linear_decision}

We interpret and analyse CLIP detectors using i) \textit{score association} to show how strongly named cues covary with the detector's logit, ii) \textit{shared-space cue similarity} to compare detector decision directions with cue directions in the cross-modal space, and iii) \textit{paired cue shifts} to analyse cue differences in content-matched real-synthetic pairs. We also report \textit{cue-span capacity} to assess how much predictive information might be lost by restricting the analysis on the cues.

\textbf{Score association}. This uses the canonical detector's logit directly. Sample $i$'s score on cue $q$ is its projection onto the cue direction of Equation~\ref{eq:cue_direction},
\begin{equation}
c_q^{(i)} = \bigl\langle \mathbf{e}^{(i)},\, \mathbf{v}_q \bigr\rangle,
\label{eq:cue_score}
\end{equation}
so positive values point toward the positive prompt pole. Score association is
\begin{equation}
r_q^{\mathrm{within}} = \tfrac{1}{2}\sum_{y \in \{0,1\}} \operatorname{Corr}_{\mathrm{pearson}}\bigl(z^{(i)},\, c_q^{(i)} \mid y^{(i)} = y\bigr),
\label{eq:score_cue_corr}
\end{equation}
each correlation over the samples of that class. A positive value means that, averaged over the two subsets, larger cue scores associate with larger synthetic-class logits; near zero means no consistent within-class covariation. Averaging within-class correlations prevents class-mean separation alone from producing a high association. Additional diagnostics and uncertainty details are reported in Appendix~\ref{appendix:score_alignment}.

\textbf{Shared-space cue similarity}. The canonical detector $D_h$ operates on the pre-projection visual representation which cannot be directly compared to cue embeddings. Therefore, we train a shared-space analysis head $D_e$  which uses the same matched training recipe

\begin{equation}
z_e^{(i)} = \mathbf{w}_e^\top {\mathbf{e}}^{(i)} + b_e,
\label{eq:projected_analysis_head}
\end{equation}
$D_e$ is not the benchmark detector, it exists only because a $1024$-d pre-projection direction cannot be compared directly with $768$-d text directions.

 \textbf{Cue-span capacity.} Sample $i$'s cue scores $\mathbf{c}^{(i)} = \bigl(c_1^{(i)}, \ldots, c_{168}^{(i)}\bigr)$ feed a classifier restricted to the vocabulary ( $D_{\mathrm{cue}}$),
\begin{equation}
z_{\mathrm{cue}}^{(i)} = \mathbf{u}^{\top} \mathbf{c}^{(i)} + b_\mathrm{cue},
\label{eq:cue_restricted}
\end{equation}
with coefficient vector $\mathbf{u}$ and intercept $b_\mathrm{cue}$. Its AUROC measures the label-predictive information accessible from the named cue scores, not agreement with the detector. The cue-restricted probe $D_{\mathrm{cue}}$ uses the same recipe on the $168$ cue scores themselves. Taken together, $D_h \rightarrow D_e \rightarrow D_{\mathrm{cue}}$ form a nested restriction: the first is the detector, the second asks how much remains after projection into CLIP's shared space, and the third asks how much information remains under the vocabulary restriction.

\textbf{Paired cue shifts.} The paired datasets let us compare each real image $\RealImage^{(i)}$ with synthetic images $\SyntheticImage^{(i,g)}$, $g$ indexing the generator, from the same source photograph or caption. These give descriptive within-pair changes in cue scores while approximately controlling content (see Appendix~\ref{appendix:paired_analysis} for details).

Cue results depend on the vocabulary and how much of the representation it covers. We read them as descriptive profiles rather than causal explanations, emphasizing recurring cue families over single directions.

Figure~\ref{fig:intro} summarizes the detector and analysis hierarchy. The same optimizer, loss, and schedule are used for all three heads.

\subsection{Concept Modeling}
\label{sec:method:concept_classifier}

Our concept model follows the Sparse Linear Concept Discovery Models (CDMs)~\cite{panousisSparseLinearConcept2023}, which use CLIP image–text similarities and a variational Bernoulli prior to obtain sparse linear concept activations. We adopt this framework for binary SID on \cnnspot{}, \synthbp{}, and \synthclic{} using the two fixed vocabularies of Section~\ref{sec:method:vocabularies}.

Let $\ConceptsDim$ be the row-normalized concept embeddings with $n$ terms, $\ImagesProjectedDim$ the row-normalized image embeddings for $m$ samples, and $\CosineSimFullDim$ their cosine similarities. Following \cite{panousisSparseLinearConcept2023}, we use a binary relevance mask $\mathbf{Z} \in\{0,1\}^{m\times n}$ to flexibly select concepts per sample. $W_c \in\mathbb{R}^{n\times 1}$ contains the classifier weights that map the gated concept similarities to the synthetic-class logit; the per-sample selection is performed by $\mathbf{Z}$. The class logits $\mathbf{a} \in\mathbb{R}^{m\times 1}$ are thus:
\begin{align}
\mathbf{a}=(\mathbf{S} \odot \mathbf{Z}) \,W_c 
\end{align}

During training, we treat the relevance mask $\mathbf{Z}$ as a latent variable and infer it with a factorized Bernoulli variational posterior. For each sample $i$ (row $\mathbf{z}_i$ of $\mathbf{Z}$), we set
\begin{align}
q(\mathbf{z}_i)
&= \text{Bernoulli}\!\left(\sigma\!\left(\ImagesProjected_i \mathbf{W}_s^\top\right)\right), \label{eq:qzi}
\\
p(\mathbf{z}_i)
&= \text{Bernoulli}(\alpha), \label{eq:pzi}
\end{align}
where $\mathbf{W}_s \in \mathbb{R}^{n \times p}$ is a learned projection from image-embedding space to per-concept logits (with $p$ the embedding dimension), $\sigma(\cdot)$ is applied element-wise, and the scalar $\alpha$ is fixed across concepts to encourage sparsity. We optimize the negative ELBO:
\begin{align}
\mathcal{L} = \mathbb{E}_{(\mathbf{x}, \mathbf{y})} \left[ \mathrm{BCEWithLogits}(\mathbf{a},\mathbf{y}) + \beta\,\mathrm{KL}\bigl(q(\mathbf{z})\,\|\,p(\mathbf{z})\bigr)\right]
\label{eq:concept_discovery_loss}
\end{align}
where the first term is the binary cross-entropy applied to the logits, $\mathrm{KL}$ regularizes the inferred masks toward the sparse prior, and $\beta$ controls the sparsity--accuracy trade-off.

We use the concept model as a complementary, intrinsically text-grounded SID classifier rather than as an explanation of the canonical linear detector: it is trained directly on class labels and is not fitted to reproduce the linear classifier's scores. Its concept contributions show how this model uses the supplied vocabulary to predict the SID label. We therefore evaluate both its detection performance and its sensitivity to the vocabulary and the sparsity weight $\beta$, and we report how stable its concept selection is across random seeds (Appendix~\ref{appendix:concept_model:selection_stability}).

\section{Experiments and Results}
\label{sec:results}

We first characterize detection performance (Section~\ref{sec:results:detection_performance}) and establish which explanation unit is stable enough to interpret (Section~\ref{sec:results:interpretation_target}). The remaining subsections then analyze what the evaluated detectors' cues are, how they differ between real and synthetic images, why cross-family generalization fails, and how the separately trained sparse concept model behaves.

\subsection{Detection Performance under Distribution Shift}
\label{sec:results:detection_performance}

We report and compare detection performance of our CLIP-based detectors with a CNN-based forensic detector, trained using the protocol described in \cite{wangCNNgeneratedImagesAre2020}. Additionally, we use a more recent detector (CommunityForensics-384 \cite{park_community_2025}), trained on approximately 5.4M images: 2.7M synthetic images generated by 4{,}803 models and a corresponding set of 2.7M real images. We evaluate the models on the test splits shown in Table~\ref{tab:data:sizes} and \textit{CF-Eval} from \cite{park_community_2025}. Results are shown in Table~\ref{tab:detector_comparison}.

\begin{table}[!ht]
\centering
\scriptsize
\begin{tabular}{llrrrr}
\toprule
Detector & Train & SynthCLIC & SynthBuster+ & CNNSpot & CF-Eval \\
\midrule
\multirow{4}{*}{CLIP (linear head)}
& SynthCLIC & 0.92 & 0.79 & 0.42 & 0.72 \\
& SynthBuster+ & 0.61 & 1.00 & 0.67 & 0.79 \\
& CNNSpot & 0.52 & 0.50 & 0.96 & 0.66 \\
& Combined & 0.87 & 0.98 & 0.88 & 0.93 \\
\midrule
\multirow{4}{*}{Forensic CNN}
& SynthCLIC & 0.98 & 0.80 & 0.49 & 0.58 \\
& SynthBuster+ & 0.76 & 0.99 & 0.52 & 0.73 \\
& CNNSpot & 0.49 & 0.52 & 0.94 & 0.53 \\
& Combined & 0.98 & 1.00 & 0.76 & 0.80 \\
\midrule
CommunityForensics-384 & broad (off-the-shelf) & 0.87 & 0.91 & 0.97 & 0.99 \\
\bottomrule
\end{tabular}
\caption{Detectors trained on different datasets are evaluated on \synthclic{}, \synthbp{}, \cnnspot{}, and the CommunityForensics-Eval (CF-Eval) dataset. Shown is mean AP across the generators in each test set.}
\label{tab:detector_comparison}
\end{table}

Low-level forensic CNNs are highly effective when training and test distributions match (e.g.\ 0.98 mAP in-domain on \synthclic{}), but their performance drops sharply under dataset and generator-family shift. The picture is similar for the CLIP linear detector: strong in-distribution results (0.96 mAP on \cnnspot{}, 0.92 on \synthclic{}) but weak transfer, falling to 0.42 mAP from \synthclic{} to \cnnspot{}.  Training on the combined dataset improves performance on CF-Eval substantially (0.93 mAP). The broad-generator CommunityForensics-384 detector transfers well across \synthclic{}, \synthbp{}, and \cnnspot{}, indicating that diverse generator exposure improves generalization. Notably, its performance on \synthclic{} (0.87) is markedly lower than on CF-Eval (0.99), suggesting \synthclic{} is a comparatively difficult distribution even for a broadly trained detector. These results do not position our CLIP linear classifier as a state-of-the-art detector. Rather, the three families fail in different places: on CF-Eval the \synthclic{}-trained CLIP head transfers better overall and on latent-diffusion and commercial generators, whereas the ProGAN-trained forensic CNN is stronger on GANs, and the broad-exposure detector is strongest everywhere (Section~\ref{sec:results:cross_family}).
The canonical detector and the factorized $k{=}8$ head achieve similar detection performance in a matched cached-feature comparison, and the same qualitative pattern holds across the three OpenAI CLIP ViT backbones we evaluate: ViT-L/14-336, ViT-B/16, and ViT-B/32 achieve comparable in-domain mAP on \synthclic{} (Appendix~\ref{appendix:clip:backbone_ablation}). Full cross-dataset and per-generator results are reported in Appendix~\ref{appendix:clip_classification_results}.

\subsection{Interpretation Target and Stability}
\label{sec:results:interpretation_target}

Before interpreting learned directions, we first assess whether they are stable across refits. We compare decision directions using the covariance-weighted cosine $\cos_\Sigma$ (Equation~\ref{eq:sigma_cosine}) and their $168$-dimensional cue-association profiles using Spearman correlation $\rho$ (Appendix~\ref{appendix:score_alignment}).

Across five seeds, the canonical detector $D_h$ is highly stable: it reaches AUROC $0.921$ on the \synthclic{} test split, with $\cos_\Sigma = 0.989$ between refit directions and $\rho = 0.991$ between their cue-association profiles. A matched factorized $k{=}8$ head achieves similar detection performance (AUROC $0.915$ versus $0.921$) and its effective direction is likewise stable ($\cos_\Sigma = 0.963$, $\rho = 0.979$). In contrast, the individual factorized axes are unstable: after Hungarian matching between refits, their mean absolute cosine is $0.289$ and their cue-profile correlation is $\rho = 0.119$. We therefore interpret the effective decision boundary rather than individual axes of the factorization (Appendix~\ref{appendix:clip:interpretation_stability}).

\begin{figure}[t]
\centering
\includegraphics[width=\linewidth]{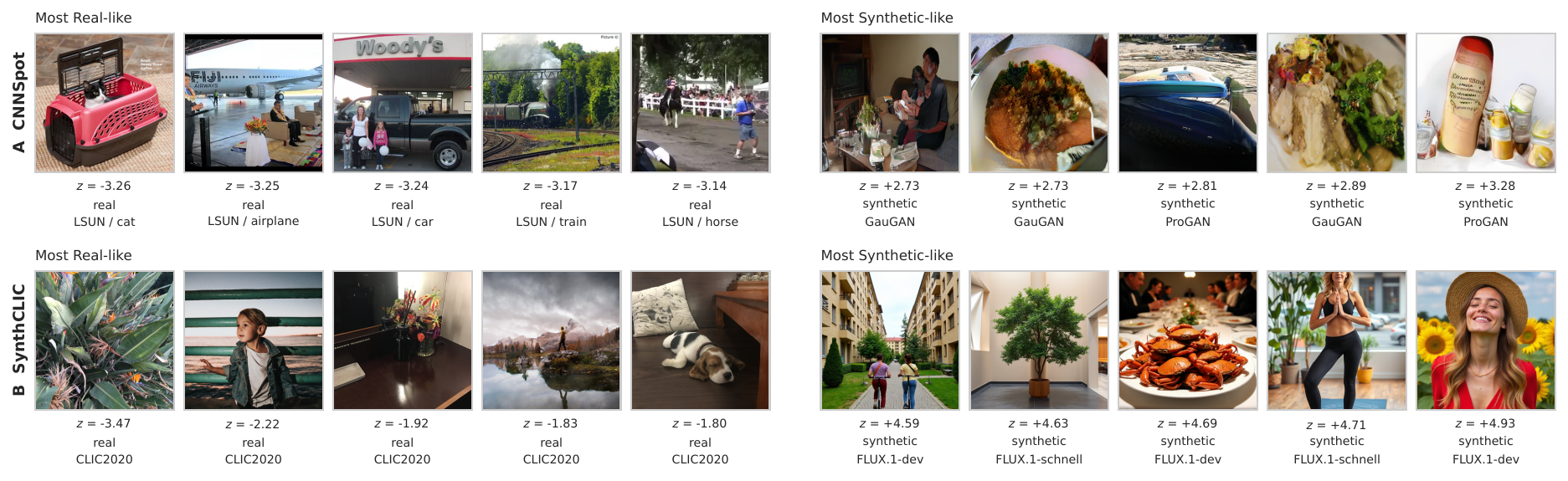}
\caption{Test-set images with the five lowest and five highest canonical-detector logits $z_h$ for \cnnspot{} (A) and \synthclic{} (B). Beneath each image the logit, the class and the image's source are shown.}
\label{fig:res:extreme_scores}
\end{figure}

Figure~\ref{fig:res:extreme_scores} illustrates the images ranked most strongly along the canonical decision direction. The extremes differ substantially between datasets: the most synthetic-scoring \cnnspot{} images are low-resolution GAN images with visible artifacts, whereas the corresponding \synthclic{} images are high-quality diffusion-generated photographs. Because the exact extreme samples vary more across refits than the decision direction itself, the montage is intended as a qualitative illustration rather than a stable characterization of individual images.

\subsection{How Much Predictive Information Is Captured by Named Cues?}
\label{sec:results:named_cues}
We next assess how much SID-relevant predictive information the fixed photography-oriented antonym vocabulary captures, by following the nested restriction $D_h \rightarrow D_e \rightarrow D_{\mathrm{cue}}$ on the \synthclic{} test split. We report AUROC here because it compares the ranking of matched heads on the same imbalanced split independently of class prevalence. Table~\ref{tab:rev:cascade} reports each step as a signed change relative to the row above. Projecting into CLIP's shared image--text space, where textual cues are defined, moves AUROC from $0.921$ for the canonical detector $D_h$ to $0.893$ for the shared-space analysis head $D_e$, a change of $-0.029$ ($95\%$ CI $[-0.037,-0.021]$). Projecting to a cue-restricted space with $D_{\mathrm{cue}}$ gives an AUROC of $0.834$, a further $-0.059$ ($95\%$ CI $[-0.075,-0.043]$). The named cue span therefore carries a substantial but incomplete portion of the linearly accessible SID signal, and most of what is lost is lost to the vocabulary rather than to the projection. This measures what the named vocabulary \emph{can predict}, not what the detector uses.

\begin{table}[!ht]
\centering
\scriptsize
\begin{tabular}{lrl}
\toprule
Stage & AUROC & $\Delta$ vs.\ previous row [95\% CI] \\
\midrule
$D_h$ (full pooler, 1024-d) & 0.921 & -- \\
$D_e$ (shared space, 768-d) & 0.893 & $-0.029$ [$-0.037$, $-0.021$] \\
$D_{\mathrm{cue}}$ (168 named cues) & 0.834 & $-0.059$ [$-0.075$, $-0.043$] \\
\bottomrule
\end{tabular}
\caption{Successive information restriction on the SynthCLIC test split (mean over 5 seeds). Each $\Delta$ is the change in AUROC relative to the row above, with a 95\% cluster bootstrap confidence interval (2000 resamples, clustered on source photo); negative values are losses. Projection into the shared image--text space costs little, restriction to the named cue span costs roughly twice as much.}
\label{tab:rev:cascade}
\end{table}

\subsection{Which Cues Covary with Detector Scores?}
\label{sec:results:score_alignment}

We correlate the evaluated detector's score with each textual cue score. The correlation is computed separately within real and synthetic images and then averaged, so Panel~A of Figure~\ref{fig:res:cue_population} identifies cues that track detector confidence among images of the same class. It does not identify the largest differences between real and synthetic images, and it does not establish that the detector causally uses a cue.
Appendix~\ref{appendix:score_alignment} reports secondary diagnostics and uncertainty details.

Higher synthetic-class scores covary with clean framing ($|r_q^{\mathrm{within}}| = 0.333$), clean low-light rendering ($0.306$), high clarity ($0.257$) and smooth long-exposure motion ($0.235$). Lower scores covary with capture-provenance cues: an instant-film texture ($0.302$), a documentary look ($0.216$) and legible text ($0.200$). The profile is distributed across many overlapping cue directions rather than carried by any one of them, so no individual cue is a complete description of the score.

\subsection{Which Cues Change Between Real and Synthetic Images?}
\label{sec:results:paired_cues}

\begin{figure}[t]
\centering
\includegraphics[width=0.5\linewidth]{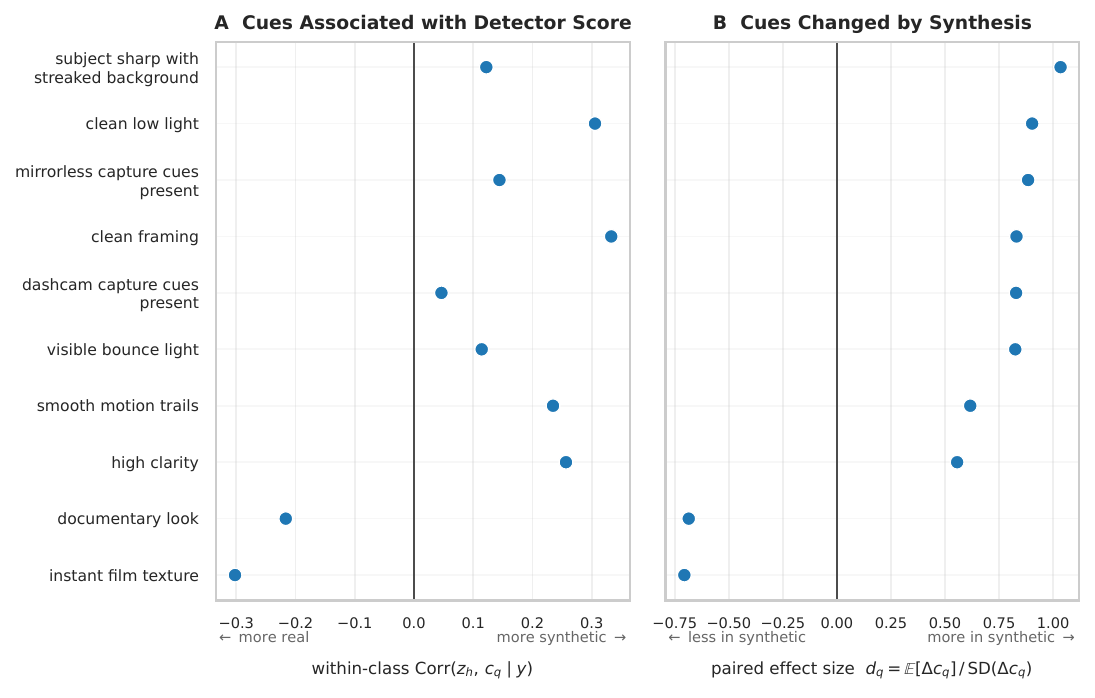}
\caption{Population-level cue interpretation on the \synthclic{} test split. \textbf{(A)} Within-class association between each named cue and the canonical detector's logit. \textbf{(B)} Paired effect size of synthesis, $d_q = \mathbb{E}[\Delta c_q]/\mathrm{SD}(\Delta c_q)$, where $\Delta c_q$ is the change in cue score between a synthetic image and its own real reference. Rows carry each cue's positive phrase and share a $y$-axis, and each panel's poles are annotated, so a row reads across both panels.}
\label{fig:res:cue_population}
\end{figure}

The paired design lets us measure cue changes while approximately
controlling image content. Panel~B of Figure~\ref{fig:res:cue_population}
reports paired effect sizes over the $1{,}712$ \synthclic{} test pairs
(Appendix~\ref{appendix:paired_analysis}). Synthetic variants shift toward
streaked-background panning ($d_q=+1.03$), cleaner low-light rendering
($+0.90$) and framing ($+0.83$), and away from instant-film texture
($-0.71$) and a documentary look ($-0.69$). Together, these distributed
changes indicate a shift toward a more polished, camera-ideal photographic
appearance rather than a single dominant artifact.

\begin{figure}[t]
\centering
\includegraphics[width=\linewidth]{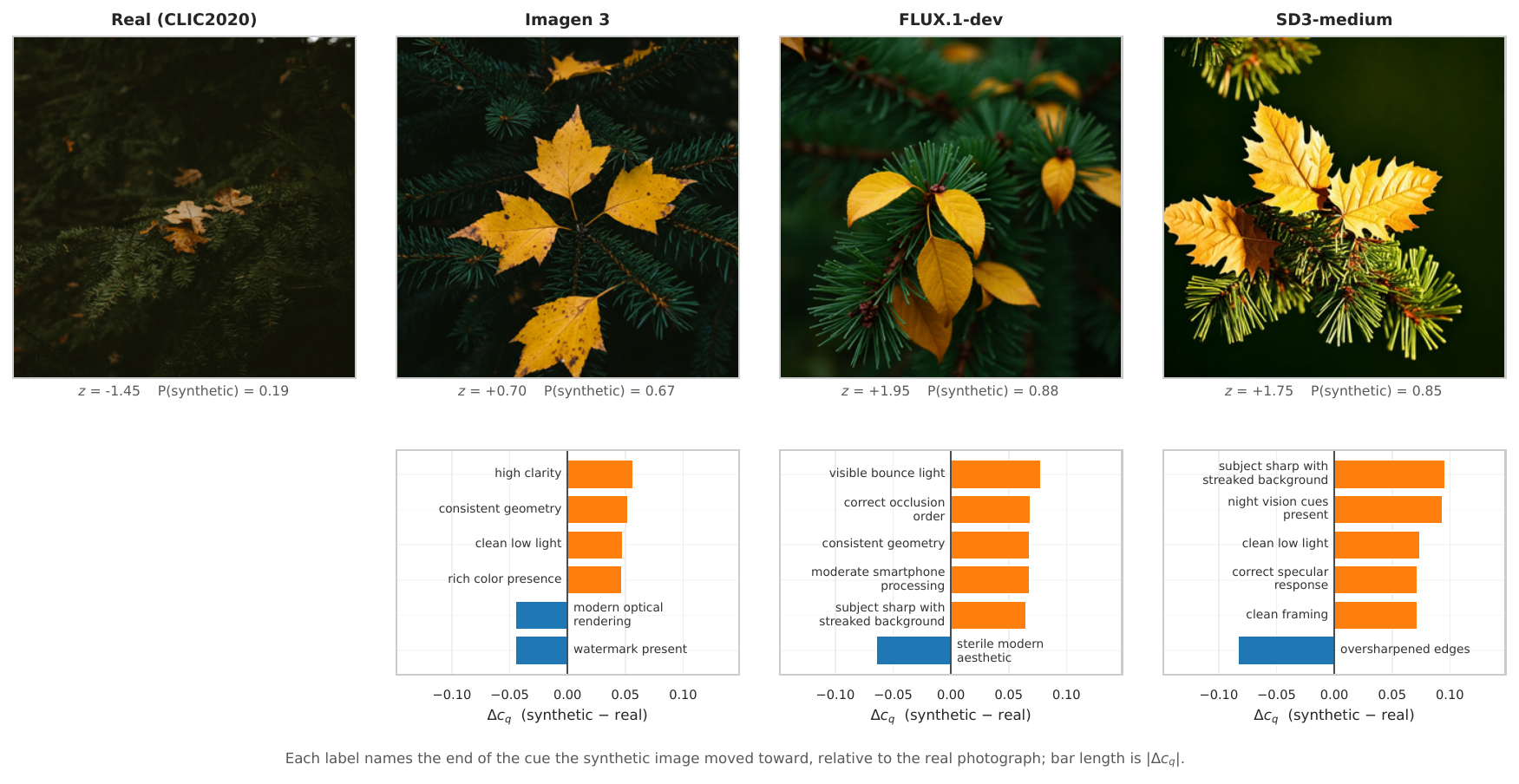}
\caption{
One CLIC photograph and three caption-conditioned synthetic counterparts. Bars show the largest cue-score changes
$\Delta c_q = c_q(x_{\text{synth}}) - c_q(x_{\text{real}})$ relative to the real reference. Positive and negative bars indicate movement toward the corresponding named cue pole. Canonical-detector logits are shown below each image. FLUX.1-schnell not shown for legibility.}
\label{fig:res:paired_example}
\end{figure}

Figure~\ref{fig:res:paired_example} illustrates these cue shifts for
one real photograph and its caption-conditioned synthetic counterparts.

\subsection{Why Does Cross-Family Generalization Fail?}
\label{sec:results:cross_family}

The weakest transfer occurs when the \synthclic{}-trained detector is
applied to GAN-heavy data. On CF-Eval, its family-level mean AP is lowest for
GAN ($0.36$) and pixel-space diffusion ($0.41$) generators, compared with
$0.87$ for latent diffusion and $0.73$ for commercial generators
(see Table~\ref{tab:rev:cfeval_architecture}). The two weakest families contain
only two and one generators, respectively, but the pattern is consistent
with the detector's low mAP on \cnnspot{} ($0.42$).

\begin{figure}[t]
\centering
\includegraphics[width=0.5\linewidth]{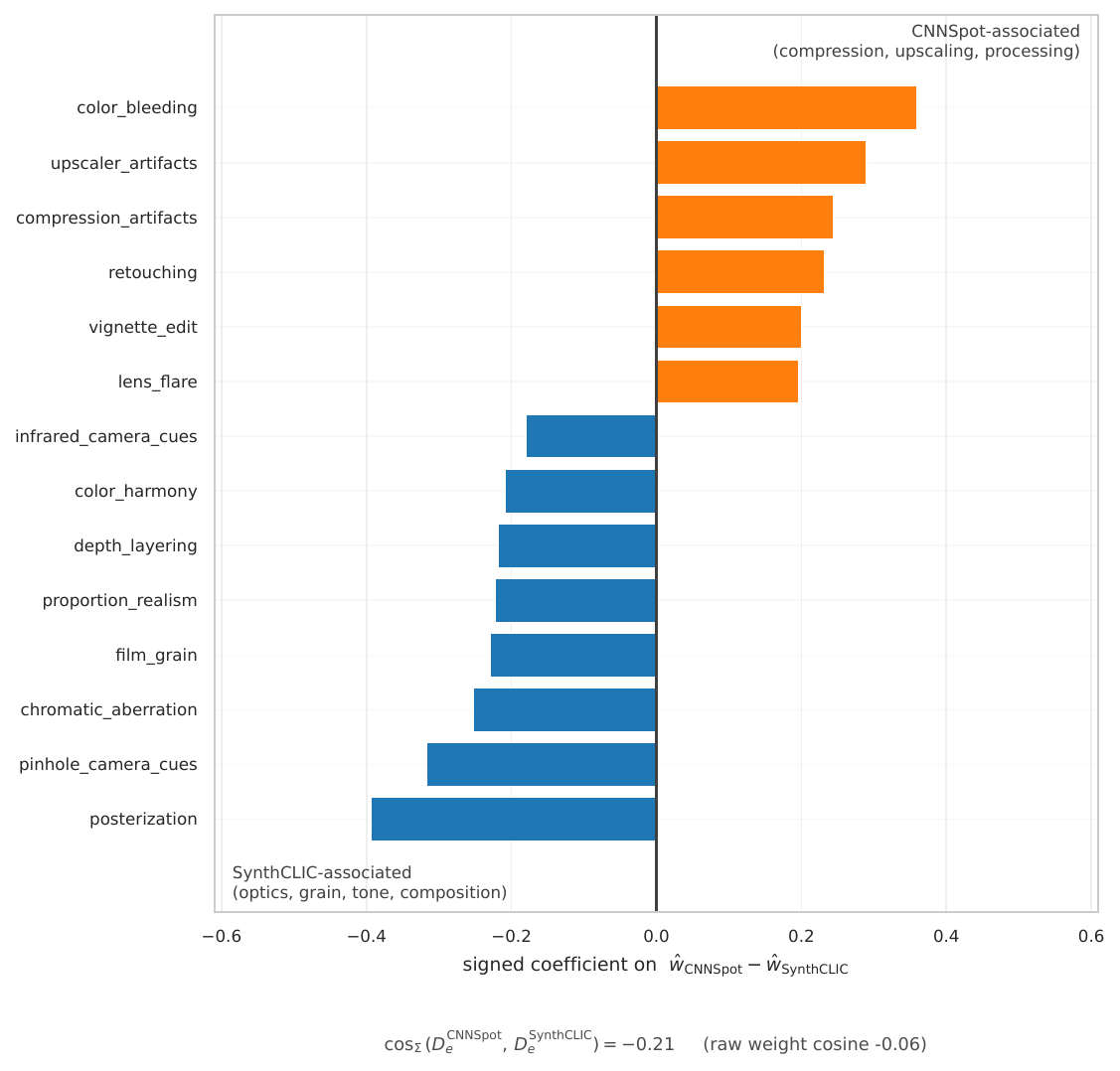}
\caption{Signed decomposition of the difference between the
\cnnspot{}- and \synthclic{}-trained shared-space decision directions onto
the named cue axes ($14$ largest coefficients; all selected across bootstrap
refits). Positive values indicate relatively stronger synthetic-class
weighting by the \cnnspot{} boundary. Processing and compression cues
dominate the \cnnspot{} side, whereas photographic, optical, and
composition-related cues dominate the \synthclic{} side.}
\label{fig:res:boundary_delta}
\end{figure}

The learned boundaries provide a geometric explanation. The
\synthclic{}- and \cnnspot{}-trained shared-space heads have a
covariance-weighted cosine of $\cos_\Sigma=-0.207$. Their signed difference
decomposes into processing- and artifact-related cues on the \cnnspot{} side
and photographic, optical, and provenance-related cues on the \synthclic{}
side (see Figure~\ref{fig:res:boundary_delta}). Thus, the two training
distributions induce substantially different decision rules rather than
different strengths of one common synthetic-image signature.

Score association gives a consistent result: on \cnnspot{}, the fixed
\synthclic{} detector retains a moderately similar cue profile
(Spearman $\rho=0.638$) even though its label ranking reverses (AUROC $0.094$), while
the independently \cnnspot{}- and \synthclic{}-trained cue profiles are
essentially unrelated ($\rho=0.008$). Cross-family failure therefore reflects
distribution-specific decision boundaries rather than uniformly weak CLIP
features, motivating broad generator exposure and complementary forensic
cues.

\begin{table}[!ht]
\centering
\scriptsize
\begin{tabular}{lrrrrrr}
\toprule
detector & mAP & GAN & LatDiff & PixDiff & Commercial & Other \\
\midrule
CLIP linear probe (SynthCLIC) & 0.72 & 0.36 & 0.87 & 0.41 & 0.73 & 0.98 \\
Forensic CNN (SynthCLIC) & 0.58 & 0.43 & 0.64 & 0.50 & 0.58 & 0.73 \\
Forensic CNN (ProGAN/CNNSpot) & 0.53 & 0.80 & 0.39 & 0.59 & 0.57 & 0.44 \\
CommunityForensics-384 (out-of-the-box) & 0.99 & 1.00 & 1.00 & 0.93 & 0.99 & 1.00 \\
\bottomrule
\end{tabular}
\caption{Per-architecture mean AP on CommunityForensics-Eval. Each value is the unweighted mean of per-generator AP within the family; mAP is the unweighted mean across all 21 generators. Family sizes are strongly unbalanced: Commercial 11, LatDiff 6, GAN 2, PixDiff 1, Other 1.}
\label{tab:rev:cfeval_architecture}
\end{table}

\subsection{The Sparse Concept Model}
\label{sec:results:concept_model}

The sparse concept model is an independently trained, text-grounded SID classifier rather than an approximation of the linear detector. It therefore asks how well the supplied vocabulary supports detection on its own.

Table~\ref{tab:res:concept_model_test_results} shows strong in-dataset performance but substantially weaker cross-dataset transfer. Compared with the linear CLIP detector (Table~\ref{tab:detector_comparison}), performance is competitive overall, with the largest drop on \synthclic{}; neither vocabulary is consistently superior.

\begin{table}[htbp]
\centering
\scriptsize
\begin{tabular}{llllll}
\toprule
& & \multicolumn{4}{c}{\textbf{Training Set}} \\
\midrule
\textbf{Test Set} & \textbf{Vocabulary} & CNNSpot & SynthBuster+ & SynthCLIC & Combined \\
\midrule
CNNSpot & Antonyms & 0.96 & 0.62 & 0.47 & 0.90 \\
 & TextSpan & 0.96 & 0.64 & 0.40 & 0.89 \\
\midrule
SynthBuster+ & Antonyms & 0.53 & 0.99 & 0.79 & 0.99 \\
 & TextSpan & 0.51 & 0.99 & 0.78 & 0.99 \\
\midrule
SynthCLIC & Antonyms & 0.54 & 0.48 & 0.85 & 0.87 \\
 & TextSpan & 0.53 & 0.53 & 0.88 & 0.88 \\
\bottomrule
\end{tabular}
\caption{mAP of sparse concept models across training sets, test sets, and vocabularies.}
\label{tab:res:concept_model_test_results}
\end{table}

\begin{figure}[htbp]
    \centering
    \includegraphics[width=0.5\linewidth]{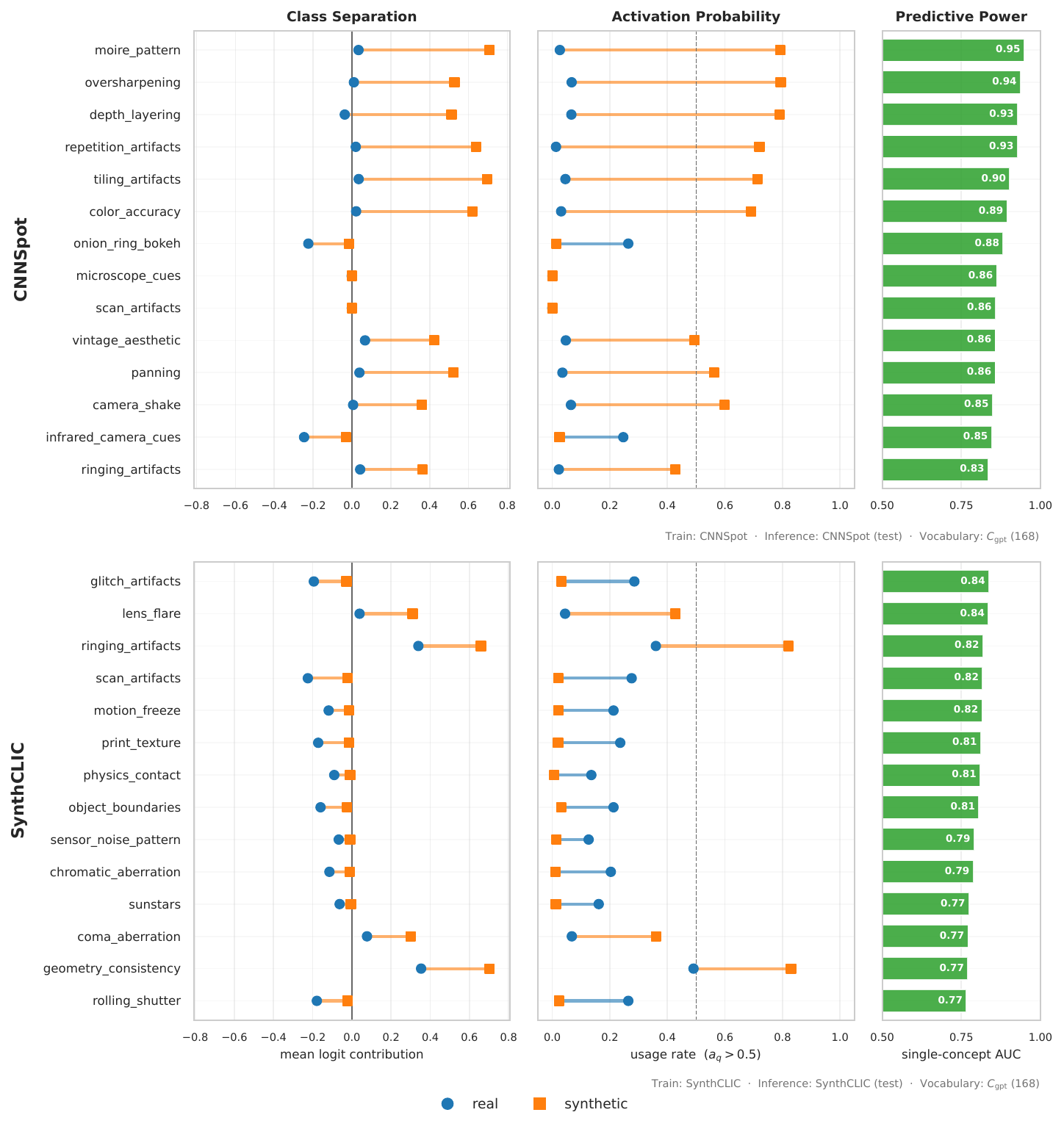}
    \caption{Representative concept profiles for models trained and evaluated on \cnnspot{} (top) and \synthclic{} (bottom). Shown are the $14$ concepts with highest single-concept AUROC, together with class-wise mean logit contribution, usage rate (gate probability $>0.5$), and single-concept AUROC. Concepts are from the antonym vocabulary $\chatgpt$.}
\label{fig:res:concept_summary}
\end{figure}

Figure~\ref{fig:res:concept_summary} shows representative concept profiles for two trained models (\synthclic{} and \cnnspot{}). The highest-ranked concepts vary across random seeds (Appendix~\ref{appendix:concept_model:selection_stability}), so we interpret recurring cue families rather than individual terms. On \cnnspot{}, the strongest concepts are synthetic-side artifact cues such as \textit{moire patterns} and \textit{tiling artifacts}. On \synthclic{}, the top concepts are less discriminative and often correspond to capture artifacts that occur more frequently in real images, such as \textit{glitch artifacts}, \textit{scan artifacts}, and \textit{print texture}. This suggests a weaker and more dataset-specific basis for classification than on \cnnspot{}.

Increasing the sparsity weight $\beta$ reduces the mean number of active concepts from about $73$ to nearly zero, while mAP varies only moderately and peaks at $\beta=10^{-4}$; the useful range is approximately $10^{-4}$--$3\times10^{-4}$ (Appendix~\ref{appendix:concept_model:beta_sensitivity}). At stronger regularization, larger remaining weights partly compensate for fewer active gates, so active-concept count is only an approximate measure of explanation complexity.

\section{Discussion \& Conclusion}\label{sec:discussion}

We revisit the three research questions formulated in Section~\ref{sec:intro}.

\textbf{Q1: How do CLIP-based SID detectors behave across paired diffusion datasets, GAN-heavy benchmarks, and cross-family evaluation settings?}
Our experiments show that CLIP-based detectors perform well when evaluated in-distribution. Cross-dataset evaluations reveal that detectors trained on \cnnspot{} generalize poorly to \synthbp{} and \synthclic{}, and vice versa, while pooled training on multiple datasets improves but does not fully close the gap. On the external CF-Eval set the same pattern resolves by generator family: the \synthclic{}-trained detector remains strong on latent-diffusion and commercial generators but is weak on the small GAN and pixel-space diffusion subsets, while a broad-exposure detector transfers across all families. These results indicate that a frozen CLIP representation does not by itself yield a universally generalizing SID detector: the discriminative structure in CLIP space depends strongly on both the real-image distribution and the family of generative models seen during training.\\

\textbf{Q2: Which explanation units for CLIP-based SID are stable and interpretable: the canonical pre-projection detector, the matched shared-space analysis head, the cue-restricted probe, or sparse text-grounded concepts?}
The actual canonical detector $D_h$ is stable: its boundary and cue-association profile are reproducible across refits, while the factorized $k{=}8$ head has a stable effective direction but unstable individual axes. The shared-space analysis head $D_e$ carries a small but measurable cost relative to $D_h$ ($+0.029$ AUROC, $95\%$ CI $[+0.021,+0.037]$), and the cue-restricted probe $D_{\mathrm{cue}}$ loses another $0.059$ AUROC relative to $D_e$, so the information available in the named vocabulary is substantial but incomplete. Sparse concept selections are also unstable in identity across seeds and vocabularies, so the exact concept list should not be treated as a unique decomposition of the detector.\\

\textbf{Q3: Which photographic, perceptual, and forensic cue families in CLIP space distinguish real from synthetic images, and how do these cues change across datasets and generator families?}
Our analyses suggest that CLIP-based SID decisions are associated with a distributed mixture of photographic, processing, and perceptual cues, and that the relevant cue profile is distribution-specific rather than a universal synthetic-image signature. Measured directly against the evaluated detector, higher synthetic scores track composition, craft and technical execution, whereas lower scores track the messiness of real capture. The signed cross-dataset decomposition shows that diffusion-era boundaries emphasize photographic, optical, and grain-related cues, whereas GAN-heavy boundaries emphasize processing and artifact cues. The two families are therefore separated by more than just a single low-level fingerprint. Content-controlled comparisons shift toward a more polished, camera-ideal photographic appearance, including cleaner framing and low-light rendering and weaker documentary/instant-film cues. CLIP-IQA attributes show systematic shifts in perceived \textit{quality}, \textit{noisiness}, and \textit{sharpness}---perceptual attributes that are not reducible to a single low-level forensic statistic. On older GAN-based data, these cues often coincide with visible artifacts and compositional issues, whereas on high-quality diffusion images they manifest as more subtle stylistic and aesthetic differences. The exact sparse concept identities remain unstable, but the aggregate evidence is robust: CLIP-based SID is associated with distributed cue families whose relative emphasis shifts between generator families and datasets.\\

\textbf{Limitations and future work.}
Our study is limited to three training datasets and one broader external evaluation benchmark, all dominated by photographic imagery, and to OpenAI CLIP variants sharing the same broad pre-training approach (ViT-L/14-336, ViT-B/16, ViT-B/32; Appendix~\ref{appendix:clip:backbone_ablation}). The cross-family decomposition is informative but not definitive, especially for the GAN and pixel-space diffusion families, which are represented by only two and one generators in CF-Eval. The canonical detector ($D_h$) exceeds the shared-space analysis head ($D_e$) by 0.029 AUROC, and the cue-restricted probe $D_{\text{cue}}$ is $0.059$ below $D_e$, so the named vocabulary remains incomplete. Sparse concept identities are stable only at the aggregate level, not in the exact concept lists selected. We therefore present the conclusions as distribution-specific and model-family-dependent rather than as a universal explanation of synthetic imagery. As a result, our conclusions may not fully transfer to non-photographic domains (e.g., medical or scientific imaging) or to future generators with qualitatively different failure modes. Our analysis focuses on photographic cues and does not explicitly combine them with low-level fingerprint-based features. Future work could explore hybrid detectors that jointly exploit semantic CLIP embeddings and forensic cues, extend our analysis to additional backbones and modalities, and investigate continual or online adaptation as new generators appear.\\

In summary, our results show that CLIP-based methods are powerful but not universal tools for synthetic image detection. They perform strongly on the older GAN-heavy \cnnspot{} benchmark, struggle more with high-quality diffusion images and cross-dataset generalization, and their decisions reflect a distributed mixture of photographic and perceptual cues---not reducible to a single stable generator fingerprint---whose relative importance shifts across datasets, making CLIP-based and low-level forensic detectors complementary. Text-grounded concept models provide useful human-readable summaries, but their interpretation depends on the vocabulary, the sparsity regularization, and the random initialization, and should not be treated as a unique or controllable decomposition of the detector. We hope that our datasets, models, and analyses provide a useful basis for developing more robust and interpretable SID systems that can keep pace with rapidly evolving generative models.

\section{Data \& Code Availability}\label{sec:data}

We will release \synthclic{}, \synthbp{}, the images used in \cnnspot{}, our antonym vocabulary $\chatgpt$, the prompts for \synthclic{} and \synthbp{}, the model checkpoints, and all inference code upon publication\footnote{\href{https://github.com/marco-willi/clip-cues}{https://github.com/marco-willi/clip-cues}}, subject to the licenses of the underlying photographs.

\section*{Declaration of generative AI and AI-assisted technologies in the manuscript preparation process}
During the preparation of this work the authors used ChatGPT to edit the manuscript and used Claude Code to create code for visualizing the results.
After using these tools, the authors reviewed and edited the content as needed and take full responsibility for the content of this work.

\bibliographystyle{plainnat}
\bibliography{bibliography}

\appendix

\section{Methodological Details}

\subsection{Datasets - \cnnspot{} Samples}
\label{appendix:datasets_example}

\begin{figure}[htbp]
    \centering
    \includegraphics[width=\linewidth]{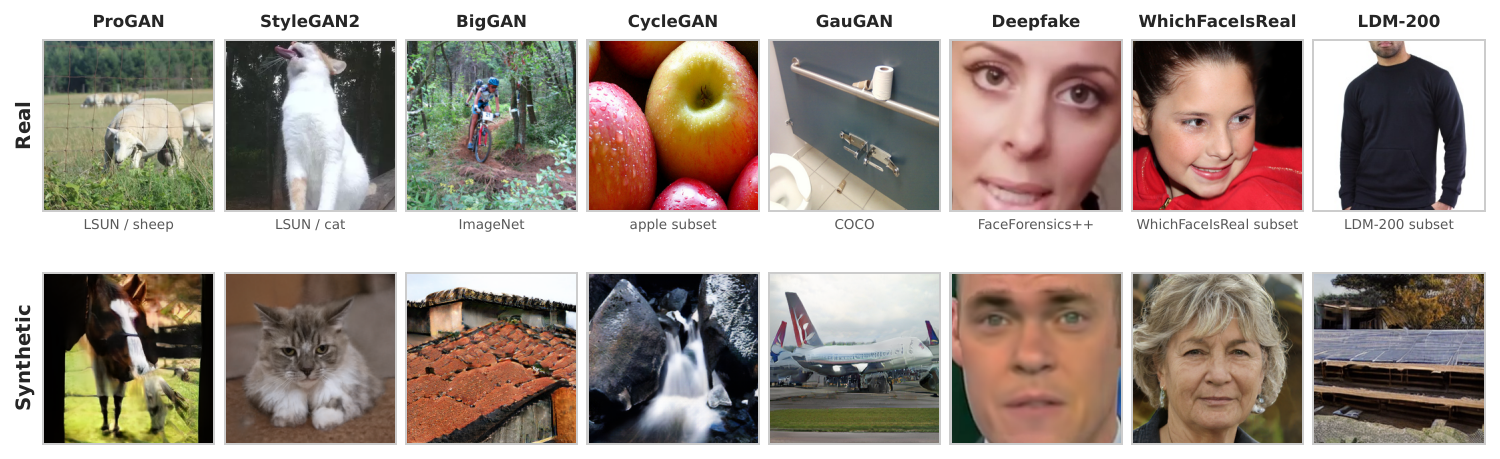}
    \caption{
    Example images from the \cnnspot{} test set, showing real and synthetic images from representative generator groups. Real images originate from the benchmark-specific source datasets indicated below each panel. Generator labels refer to the corresponding synthetic subset. Examples are selected near the median canonical-detector score within each group.}
    \label{fig:data:cnnspot}
\end{figure}

Figure \ref{fig:data:cnnspot} shows a few example images from the \cnnspot{} test set.

\subsection{Datasets - Generative Models Used}
\label{appendix:datasets_genmodels}

For the datasets \synthbp{} and \synthclic{}, we generated synthetic images and captions using the following model versions: Imagen3~\cite{imagen-team-googleImagen32024}\footnote{\textit{imagen-3.0-generate-002}}, FluxDev\footnote{\href{https://huggingface.co/black-forest-labs/FLUX.1-dev}{https://huggingface.co/black-forest-labs/FLUX.1-dev}}, FluxSchnell\footnote{\href{https://huggingface.co/black-forest-labs/FLUX.1-schnell}{https://huggingface.co/black-forest-labs/FLUX.1-schnell}}, Stable Diffusion 3 Medium\footnote{\href{https://huggingface.co/stabilityai/stable-diffusion-3-medium}{https://huggingface.co/stabilityai/stable-diffusion-3-medium}}, and Gemini\footnote{\textit{gemini-2.0-flash-001}} (for caption generation).

\subsection{CLIP Classifier - Implementation Details}
\label{appendix:method:clip_classifier_implementation_details}

Table \ref{tab:clip_classifier_hyperparameters_detailed} summarizes the canonical detector and the factorized $k{=}8$ configuration. The stability analysis instead refits both heads on identical cached, unaugmented features so that they differ only in factorization. When using the shared embedding space $\mathbf{e}$, we rescale it and train on $\tilde{\mathbf{e}}^{(i)}$. A single global scale factor equalizes the representation scales of $\mathbf{h}^{(i)}$ and $\mathbf{e}^{(i)}$, so that the fixed $\ell_2$ weight decay of the shared recipe acts as the same amount of regularization in both spaces while the angular geometry is left unchanged.

\begin{table}[ht]
\scriptsize
\centering
\caption{Implementation Details for CLIP Classifier}
\label{tab:clip_classifier_hyperparameters_detailed}
\footnotesize
\begin{tabular}{@{}lllp{3.5cm}@{}}
\toprule
\textbf{Category} & \textbf{Parameter} & \textbf{Value} & \textbf{Notes} \\
\midrule
Training & Max Epochs & 200 & With early stopping \\
 & Batch Size & 64 & \\
 & Early Stopping Patience & 5 epochs & Based on validation loss \\
 & Validation Frequency & Every epoch & \\
 & Accelerator & gpu (single) &  \\
Optimizer & Type & Adam & torch.optim.Adam \\
 & Learning Rate & $1e-3$ (default) &\\
 & Weight Decay & 0.01 & \\
 & Beta Parameters & $(0.9, 0.999)$ &  defaults \\
 & Epsilon & $1e-8$ & default \\
Model Architecture & Feature Extractor & clip-large-patch14 & openai/clip-vit-large-patch14-336 \\
 & Feature Extraction Layer & pooler output & Layer-normalized final-layer \texttt{[CLS]}, $d=1024$ \\
 & Frozen Backbone & Yes & No gradient updates to CLIP \\
Loss Function & Primary Loss & Binary Cross-Entropy with Logits &  \\
 & Label Smoothing & $y = y*(1-\epsilon) + (1-y)*\epsilon$ & $\epsilon=0.1$ \\
\midrule
\multicolumn{4}{@{}l}{\textit{Canonical linear classifier ($k{=}1$; all main-text CLIP detector-performance results)}} \\
Data Aug. (Train) & Augmentations & None & Deterministic pre-processing only (Section~\ref{sec:method:clip_linear_classifier}) \\
Model Architecture & Classification Head & single linear layer & $\WeightLinear \in \mathbb{R}^{d}$, no hidden layer \\
\midrule
\multicolumn{4}{@{}l}{\textit{Factorized diagnostic head ($k{=}8$; Appendix~\ref{appendix:clip:orthogonal_head} only)}} \\
Data Aug. (Train) & RandomResizedCrop & scale=(0.5, 1.0), size=512 & Crop 50--100\% of original, resize to 512$\times$512 (applied before CLIP pre-processing; the final input is resized to the model's resolution) \\
 & RandomHorizontalFlip & p=0.5 & \\
 & RandomJPEGCompression & quality=(65, 100) & Uniform sampling \\
Training & Precision & bf16-mixed & Mixed-precision; applies only to the factorized-head runs, which re-encode images. The canonical classifier is fitted on cached embeddings in fp32 \\
Model Architecture & Classification Head & orthogonal activation head & \\
 & Intermediate Dimensions & 8 & Single intermediate projection; no nonlinearity \\
 & Orthogonal Initialization & Yes & Applied to weights \\
Regularization & Orthogonality Loss Weight & 0.33 & Auxiliary orthogonality loss, Equation~\ref{eq:activation_orthogonality} \\
\bottomrule
\end{tabular}
\end{table}

\subsection{Factorized Orthogonal Head}
\label{appendix:clip:orthogonal_head}

As an auxiliary interpretability experiment, we consider a factorized linear head with $k=8$ dimensions. For a pre-projection visual representation $\mathbf{h}^{(i)} \in \mathbb{R}^{1024}$, the logit is
\begin{equation}
z^{(i)}
=
\mathbf{h}^{(i)\top}\mathbf{W}_1\mathbf{w}_2 + b,
\qquad
\mathbf{W}_1 \in \mathbb{R}^{1024 \times k},
\quad
\mathbf{w}_2 \in \mathbb{R}^{k}.
\end{equation}
Because no non-linearity is applied, the head is equivalent to a single linear classifier with effective direction
\begin{equation}
\mathbf{w}_{\mathrm{eff}} = \mathbf{W}_1\mathbf{w}_2.
\end{equation}

To encourage distinct intermediate activation dimensions, let
$\mathbf{A}=\mathbf{H}\mathbf{W}_1 \in \mathbb{R}^{n \times k}$
denote the mini-batch activations and let $\bar{\mathbf{A}}$ denote $\mathbf{A}$ after $\ell_2$-normalizing each column. We optimize
\begin{equation}
\mathcal{L}
=
\mathcal{L}_{\mathrm{BCE}}
+
\lambda
\left\|
\mathbf{I}
-
\bar{\mathbf{A}}^\top\bar{\mathbf{A}}
\right\|_F^2,
\qquad
\lambda = 0.33.
\label{eq:activation_orthogonality}
\end{equation}
The regularizer therefore acts directly on the activation dimensions used for interpretation.

An alternative is to enforce weight orthogonality through
$\|\mathbf{I}-\mathbf{W}_1^\top\mathbf{W}_1\|_F^2$.
However, orthogonal weights do not generally imply orthogonal activations because
\begin{equation}
\mathbf{A}^\top\mathbf{A}
=
\mathbf{W}_1^\top
\mathbf{H}^\top\mathbf{H}
\mathbf{W}_1,
\end{equation}
which also depends on the empirical feature geometry. We therefore use activation orthogonality when the intermediate dimensions themselves are the object of analysis. Appendix~\ref{appendix:clip:orthogonality_ablation} compares both regularizers empirically.

\subsection{Interpretation Stability}
\label{appendix:clip:interpretation_stability}

We assess the stability of learned linear directions across five independent refits. Because raw parameter-space cosine can be misleading when feature dimensions have unequal variance and covariance, we use the covariance-weighted cosine
\begin{equation}
\cos_{\Sigma}(\mathbf{w}_1,\mathbf{w}_2)
=
\frac{
\mathbf{w}_1^\top \mathbf{\Sigma}\mathbf{w}_2
}{
\sqrt{\mathbf{w}_1^\top \mathbf{\Sigma}\mathbf{w}_1}
\sqrt{\mathbf{w}_2^\top \mathbf{\Sigma}\mathbf{w}_2}
},
\label{eq:sigma_cosine}
\end{equation}
where $\mathbf{\Sigma}$ is the empirical feature covariance. This quantity measures agreement between the centered linear scores induced by two directions on the observed feature distribution.

For the canonical detector $D_h$ and the factorized $k{=}8$ head, $\mathbf{\Sigma}$ is computed from the \synthclic{} pre-projection visual representations. Cue-profile stability is measured by the Spearman correlation $\rho$ between the corresponding $168$-dimensional cue-association profiles. Individual $k{=}8$ axes are matched between refits with the Hungarian algorithm using absolute cosine similarity to account for permutation and sign ambiguity.

\begin{table}[ht]
\centering
\begin{tabular}{lccc}
\toprule
Explanation unit & AUROC & $\cos_\Sigma$ & Cue-profile $\rho$ \\
\midrule
Canonical detector $D_h$          & 0.921 & 0.989 & 0.991 \\
Factorized $k{=}8$: effective direction & 0.915 & 0.963 & 0.979 \\
Factorized $k{=}8$: individual axes     & --    & 0.289 & 0.119 \\
\bottomrule
\end{tabular}
\caption{Stability across five independent refits on \synthclic{}. Direction similarity is measured with the covariance-weighted cosine $\cos_\Sigma$; cue-profile similarity is measured with Spearman correlation $\rho$. For the factorized $k{=}8$ head, the effective direction is $\mathbf{w}_{\mathrm{eff}}=\mathbf{W}_1\mathbf{w}_2$. Individual axes are compared after Hungarian matching.}
\label{tab:interpretation_stability}
\end{table}

The effective decision boundary is therefore reproducible across refits, whereas the individual axes of the $k{=}8$ factorization are not.

\subsection{Concept Model - Implementation Details}

Table \ref{tab:concept_model_hyperparams} shows the implementation details of the concept model (see Section \ref{sec:method:concept_classifier}).

\begin{table}[ht]
\centering
\small
\begin{tabular}{@{}llp{5cm}@{}}
\toprule
\textbf{Category} & \textbf{Parameter} & \textbf{Value} \\
\midrule
\multirow{6}{*}{Training} & Max Epochs & 4000 \\
& Batch Size & 256 \\
& Early Stopping Patience & 10 epochs \\
& Validation Frequency & Every 40 epochs \\
& Random Seed & 123 \\
& Early Stopping Metric & val/loss \\
\midrule
\multirow{3}{*}{Optimizer} & Type & AdamW \\
& Learning Rate & $1 \times 10^{-3}$ \\
& Weight Decay & $1 \times 10^{-4}$ \\
\midrule
\multirow{4}{*}{Model} & Temperature ($\tau$) & 0.1 \\
& Sparsity Weight ($\beta$) & $1 \times 10^{-4}$ \\
& Target Sparsity ($\alpha$) & $1 \times 10^{-4}$ \\
\midrule
\multirow{1}{*}{Vocabularies} & $C$ & $\chatgpt$, $\textspan$ \\
\bottomrule
\end{tabular}
\caption{Hyperparameters for Concept Bottleneck Model Experiments}
\label{tab:concept_model_hyperparams}
\end{table}

\section{Additional Results}

\subsection{Characterizing Datasets with CLIP-IQA}
\label{appendix:clip_iqa}

\begin{figure}[htbp]
    \centering
    \includegraphics[width=\linewidth]{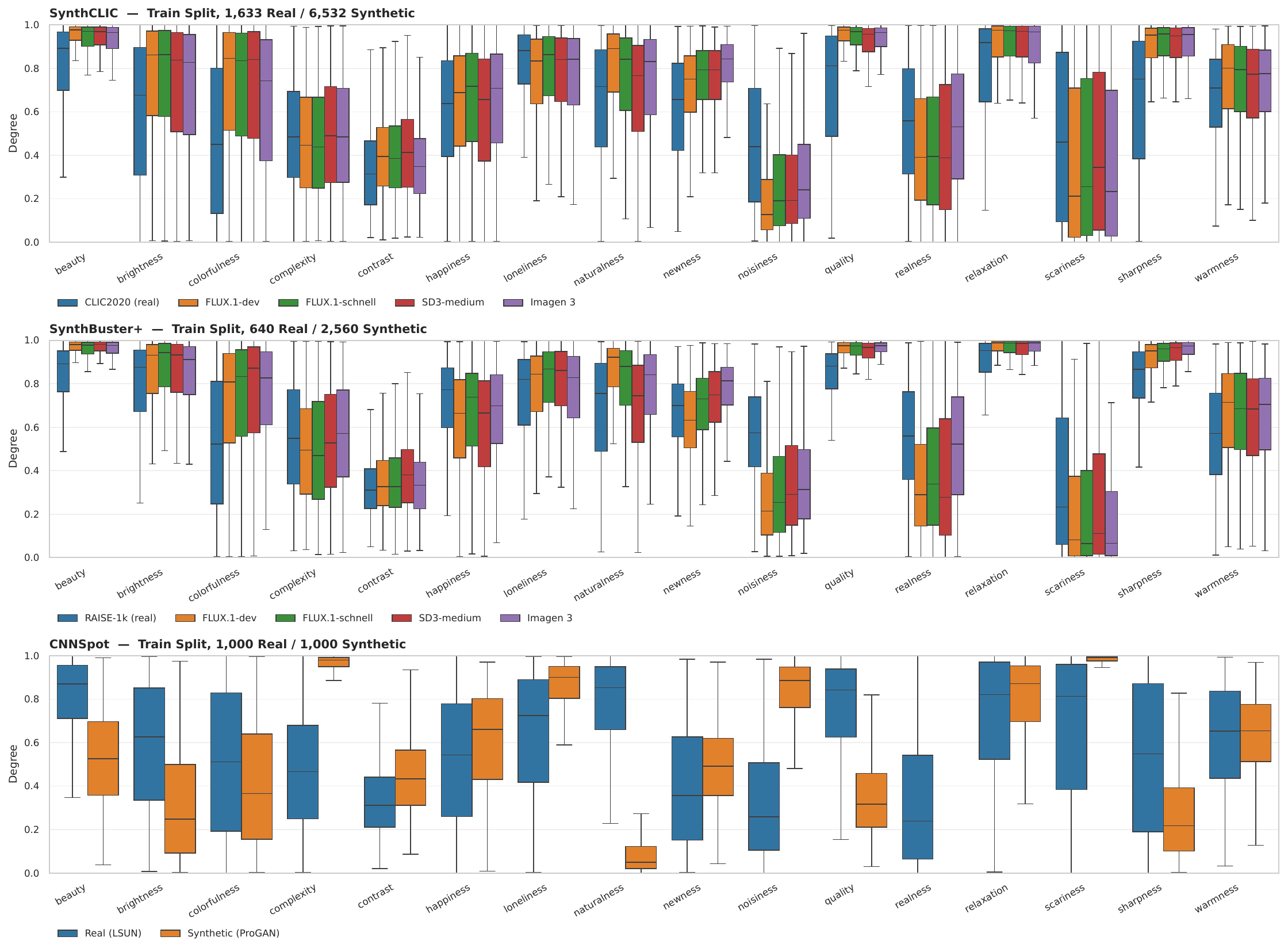}
    \caption{Distribution of 16 CLIP-IQA \cite{wangExploringCLIPAssessing2022} attribute scores for \cnnspot{}, \synthbp{}, and \synthclic{}, grouped by image source. The scores represent CLIP-based perceptual attributes rather than direct pixel measurements.}
    \label{fig:appendix:clip_iqa}
\end{figure}

We use CLIP-IQA~\cite{wangExploringCLIPAssessing2022} to compare perceptual image attribute scores across the datasets and image sources. Figure~\ref{fig:appendix:clip_iqa} shows the resulting score distributions.

\subsection{Content-Controlled Paired Analysis}
\label{appendix:paired_analysis}

The paired analysis of Section~\ref{sec:results:paired_cues} exploits the structure of \synthclic{}, where each real photograph has synthetic counterparts generated from its caption. For cue direction $\mathbf{v}_q$ and shared-space representation $\mathbf{e}$, we define the cue shift for source photograph $i$ and generator $g$ as
\begin{align}
\Delta c_{igq}
=
\bigl\langle \mathbf{e}^{(i,g)}_{\mathrm{syn}}, \mathbf{v}_q \bigr\rangle
-
\bigl\langle \mathbf{e}^{(i)}_{\mathrm{real}}, \mathbf{v}_q \bigr\rangle,
\label{eq:paired_cue_shift}
\end{align}
where positive values indicate movement toward the positive pole of the attribute. We summarize each cue by the standardized paired effect
\begin{align}
d_q
=
\frac{\mathrm{mean}_{ig}\!\left(\Delta c_{igq}\right)}
     {\mathrm{sd}_{ig}\!\left(\Delta c_{igq}\right)}.
\label{eq:paired_effect}
\end{align}

The analysis uses the \synthclic{} test split, comprising $428$ source photographs and four synthetic counterparts per photograph ($1{,}712$ pairs). Figure~\ref{fig:res:cue_population} pools all pairs when estimating $d_q$. Uncertainty in the mean cue shifts is estimated with a cluster bootstrap over source photographs ($2{,}000$ draws), keeping each photograph and its synthetic counterparts in the same resampled cluster. The pairing controls image content only approximately: synthetic counterparts are generated from captions of the real photographs and therefore remain semantically similar but are not identical in content.

\subsection{Score-Space Cue Alignment}
\label{appendix:score_alignment}

Additional diagnostics support the score-association analysis of
Section~\ref{sec:results:score_alignment}. On the \synthclic{} test split,
the median $|r_q^{\mathrm{within}}|$ across the $168$ cues is $0.119$,
compared with $0.459$ for the strongest cue, indicating a distributed
association profile rather than dominance by a single attribute. The shared-space analysis head $D_e$ produces a broadly similar cue ranking to the canonical
detector $D_h$ (Spearman $\rho = 0.864$). Uncertainty estimates use a cluster
bootstrap over source photographs so that paired variants are resampled
together.

\subsection{Decision Boundary Differences Across Training Datasets}
\label{appendix:clip:boundary_decomposition}

We fit three independent $D_e$ heads per training dataset and report
cross-dataset decision boundary similarities averaged over all nine cross-seed pairs.
We use the covariance-weighted cosine of Equation~\ref{eq:sigma_cosine}. Under the \synthclic{} shared-space covariance,
\begin{equation}
\cos_\Sigma(D_{e,\mathrm{SynthCLIC}},D_{e,\mathrm{CNNSpot}})
=
-0.2068.
\end{equation}
compared with a raw cosine of $-0.061$. \synthbp{} is positively aligned with both \synthclic{} ($\cos_\Sigma=0.311$) and \cnnspot{} ($0.353$), indicating that the strong disagreement is specific to the \synthclic{}--\cnnspot{} pair rather than a general opposition between generator families.

Figure~\ref{fig:res:boundary_delta} decomposes the signed difference between the \cnnspot{}- and \synthclic{}-trained directions onto the $168$ cue axes. The dominant terms separate processing and artifact cues on the \cnnspot{} side from photographic, optical, and grain-related cues on the \synthclic{} side, supporting a distribution-specific interpretation of the learned boundary. Because \cnnspot{} is nearly perfectly separable in the shared representation, its exact fitted normal is weakly identified; we therefore interpret the boundary similarities and cue decomposition as diagnostic summaries rather than as a unique or universal explanation of synthetic imagery.

\subsection{Concept Modeling}
\label{appendix:concept_model}

\subsubsection{Sparsity Sensitivity ($\beta$)}
\label{appendix:concept_model:beta_sensitivity}

Table~\ref{tab:rev:beta_sensitivity} shows the effect of varying the sparsity weight $\beta$ on concept-model detection and on the number of active concepts (mean $\pm$ std over five seeds). Detection peaks at $\beta = 10^{-4}$ ($0.88 \pm 0.01$ mAP with $13.7$ active concepts) and declines mildly toward both ends of the range, to $0.84$ at $\beta = 10^{-5}$ and at $\beta = 10^{-2}$, while the mean number of active concepts falls from about $73$ to $0.2$. A useful operating range is therefore roughly $10^{-4}$ to $3\times10^{-4}$, combining strong detection with moderate gate activity.

At the strongest regularization, the classifier compensates for fewer active gates with larger remaining weights, so active-concept count should not be interpreted directly as explanation complexity.

\begin{table}[!ht]
\centering
\scriptsize
\begin{tabular}{lllll}
\toprule
$\beta$ & mAP & \#active & gate mass & max$|W_{\mathrm{clf}}|$ \\
\midrule
$10^{-5}$ & 0.84 $\pm$ 0.01 & 72.6 $\pm$ 2.1 & 73.00 $\pm$ 1.98 & 6 $\pm$ 0 \\
$10^{-4}$ & 0.88 $\pm$ 0.01 & 13.7 $\pm$ 2.7 & 14.78 $\pm$ 2.75 & 13 $\pm$ 2 \\
$3\times10^{-4}$ & 0.87 $\pm$ 0.02 & 6.7 $\pm$ 1.8 & 6.98 $\pm$ 1.84 & 21 $\pm$ 6 \\
$10^{-3}$ & 0.84 $\pm$ 0.02 & 1.8 $\pm$ 0.5 & 1.94 $\pm$ 0.49 & 32 $\pm$ 7 \\
$10^{-2}$ & 0.84 $\pm$ 0.01 & 0.2 $\pm$ 0.1 & 0.29 $\pm$ 0.06 & 51 $\pm$ 11 \\
\bottomrule
\end{tabular}
\caption{Concept-sparsity weight $\beta$ vs detection on SynthCLIC (per-generator mAP), mean number of active concepts, total gate mass, and max classifier weight (mean $\pm$ std over 5 seeds). As $\beta$ grows the gates collapse but the classifier rescales (max$|W_{\mathrm{clf}}|$ grows), so mAP is preserved at near-zero active concepts.}
\label{tab:rev:beta_sensitivity}
\end{table}

\subsubsection{Concept-Selection Stability}
\label{appendix:concept_model:selection_stability}
Across seeds, detection performance and mean sparsity remain stable, but the selected concepts do not. At $\beta = 10^{-4}$, the top-30 concept sets have mean Jaccard overlap $\approx 0.24$. We therefore interpret recurring cue families rather than individual concepts and treat Figure~\ref{fig:res:concept_summary} as a representative profile rather than a unique explanation.

\subsection{CLIP Classifier}
\label{appendix:clip_classification_results}

Table \ref{tab:clip:test_results_detail} shows detailed results for each model.

\begin{table}[!ht]
\centering
\scriptsize
\setlength{\tabcolsep}{4pt}
\begin{tabular}{llrrrrrrrr}
\toprule
 & & \multicolumn{2}{c}{SynthCLIC} & \multicolumn{2}{c}{SynthBuster+} & \multicolumn{2}{c}{CNNSpot} & \multicolumn{2}{c}{Combined} \\
\cmidrule(lr){3-4} \cmidrule(lr){5-6} \cmidrule(lr){7-8} \cmidrule(lr){9-10}
Test set & Generator & ACC & AP & ACC & AP & ACC & AP & ACC & AP \\
\midrule
\multirow{5}{*}{SynthCLIC}
& FLUXDev & 0.71 & 0.93 & 0.50 & 0.58 & 0.49 & 0.47 & 0.65 & 0.87 \\
& FLUXSchnell & 0.71 & 0.95 & 0.50 & 0.61 & 0.50 & 0.53 & 0.65 & 0.91 \\
& Imagen3 & 0.70 & 0.88 & 0.50 & 0.58 & 0.49 & 0.48 & 0.64 & 0.80 \\
& SD3M & 0.71 & 0.93 & 0.50 & 0.67 & 0.52 & 0.61 & 0.66 & 0.89 \\
\cmidrule(l){2-10}
& \textit{mean} & \textit{0.71} & \textit{0.92} & \textit{0.50} & \textit{0.61} & \textit{0.50} & \textit{0.52} & \textit{0.65} & \textit{0.87} \\
\midrule
\multirow{14}{*}{SynthBuster+}
& DALLE2 & 0.39 & 0.43 & 0.88 & 0.97 & 0.65 & 0.73 & 0.76 & 0.88 \\
& DALLE3 & 0.77 & 0.98 & 0.90 & 1.00 & 0.33 & 0.31 & 0.81 & 1.00 \\
& Firefly & 0.57 & 0.63 & 0.90 & 0.99 & 0.53 & 0.57 & 0.80 & 0.97 \\
& FluxDev & 0.77 & 0.97 & 0.90 & 1.00 & 0.33 & 0.32 & 0.81 & 0.99 \\
& FluxSchnell & 0.77 & 0.98 & 0.90 & 1.00 & 0.34 & 0.33 & 0.81 & 0.99 \\
& Glide & 0.53 & 0.57 & 0.90 & 1.00 & 0.61 & 0.70 & 0.81 & 0.98 \\
& Imagen3 & 0.76 & 0.95 & 0.90 & 0.99 & 0.34 & 0.33 & 0.81 & 0.97 \\
& MJv5 & 0.75 & 0.93 & 0.90 & 0.99 & 0.36 & 0.35 & 0.81 & 0.98 \\
& SD1.3 & 0.66 & 0.77 & 0.90 & 1.00 & 0.48 & 0.52 & 0.81 & 0.99 \\
& SD1.4 & 0.69 & 0.78 & 0.90 & 1.00 & 0.48 & 0.50 & 0.81 & 0.98 \\
& SD2 & 0.50 & 0.58 & 0.90 & 1.00 & 0.65 & 0.76 & 0.81 & 0.98 \\
& SD3M & 0.77 & 0.96 & 0.90 & 1.00 & 0.40 & 0.40 & 0.81 & 0.99 \\
& SDXL & 0.64 & 0.78 & 0.90 & 1.00 & 0.55 & 0.63 & 0.81 & 0.99 \\
\cmidrule(l){2-10}
& \textit{mean} & \textit{0.66} & \textit{0.79} & \textit{0.90} & \textit{1.00} & \textit{0.47} & \textit{0.50} & \textit{0.81} & \textit{0.98} \\
\midrule
\multirow{22}{*}{CNNSpot}
& BigGAN & 0.46 & 0.47 & 0.51 & 0.66 & 0.98 & 1.00 & 0.79 & 0.96 \\
& CRN & 0.72 & 0.74 & 0.58 & 0.78 & 0.74 & 1.00 & 0.83 & 0.90 \\
& CycleGAN & 0.31 & 0.35 & 0.52 & 0.76 & 0.93 & 1.00 & 0.63 & 0.91 \\
& DALLE & 0.34 & 0.36 & 0.51 & 0.69 & 0.97 & 1.00 & 0.75 & 0.89 \\
& DeepFake & 0.55 & 0.60 & 0.51 & 0.50 & 0.69 & 0.92 & 0.65 & 0.76 \\
& GauGAN & 0.42 & 0.41 & 0.52 & 0.88 & 0.97 & 1.00 & 0.82 & 0.98 \\
& GLIDE-100-10 & 0.27 & 0.34 & 0.52 & 0.74 & 0.86 & 0.98 & 0.76 & 0.89 \\
& GLIDE-100-27 & 0.26 & 0.34 & 0.52 & 0.75 & 0.85 & 0.98 & 0.75 & 0.88 \\
& GLIDE-50-27 & 0.28 & 0.34 & 0.52 & 0.74 & 0.88 & 0.99 & 0.76 & 0.90 \\
& Guided & 0.34 & 0.36 & 0.53 & 0.67 & 0.73 & 0.90 & 0.66 & 0.72 \\
& IMLE & 0.54 & 0.55 & 0.56 & 0.75 & 0.74 & 1.00 & 0.85 & 0.92 \\
& LDM-100 & 0.24 & 0.33 & 0.52 & 0.75 & 0.98 & 1.00 & 0.77 & 0.91 \\
& LDM-200 & 0.23 & 0.33 & 0.52 & 0.76 & 0.97 & 1.00 & 0.76 & 0.91 \\
& LDM-200-CFG & 0.41 & 0.40 & 0.52 & 0.66 & 0.86 & 0.98 & 0.74 & 0.84 \\
& ProGAN & 0.25 & 0.33 & 0.50 & 0.68 & 1.00 & 1.00 & 0.98 & 1.00 \\
& SAN & 0.43 & 0.42 & 0.52 & 0.63 & 0.57 & 0.71 & 0.55 & 0.64 \\
& SeeingDark & 0.46 & 0.45 & 0.57 & 0.87 & 0.67 & 0.83 & 0.70 & 0.78 \\
& StarGAN & 0.45 & 0.44 & 0.44 & 0.33 & 0.99 & 1.00 & 0.87 & 0.92 \\
& StyleGAN & 0.39 & 0.41 & 0.53 & 0.50 & 0.88 & 0.99 & 0.92 & 0.98 \\
& StyleGAN2 & 0.47 & 0.46 & 0.50 & 0.46 & 0.77 & 0.98 & 0.87 & 0.94 \\
& WhichFaceIsReal & 0.46 & 0.45 & 0.53 & 0.48 & 0.94 & 0.99 & 0.73 & 0.86 \\
\cmidrule(l){2-10}
& \textit{mean} & \textit{0.39} & \textit{0.42} & \textit{0.52} & \textit{0.67} & \textit{0.86} & \textit{0.96} & \textit{0.77} & \textit{0.88} \\
\bottomrule
\end{tabular}
\caption{Per-generator detection performance of the CLIP linear head ($D_h$, 1024-d frozen \texttt{pooler\_output}), for each training set (columns) on each test set (rows). AP is the per-generator average precision of the paper's Convention~A: on \cnnspot{} each generator is scored against \emph{its own} real images (\texttt{matched} pairing), on \synthclic{} and \synthbp{} against the single shared real pool (\texttt{shared} pairing). ACC uses the head's own decision rule, $\hat{p} > 0.5$ (equivalently logit $z > 0$), applied uniformly to every cell and computed on the same generator group as AP; no threshold is tuned on any test set. Per-column means reproduce the cross-dataset matrix of Table~\ref{tab:detector_comparison}. All heads are trained under one matched recipe on frozen cached features without augmentation.}
\label{tab:clip:test_results_detail}
\end{table}

Table \ref{tab:k_ablation_overall} shows ablations with respect to the number $k$ of orthogonal dimensions.

\begin{table}[ht]
\centering
\scriptsize
\begin{tabular}{lcccc}
\toprule
k & 2 & 4 & 8 & 16 \\
Test Dataset &  &  &  &  \\
\midrule
CNNSpot & 0.39 & 0.38 & 0.37 & 0.38 \\
SynthBuster+ & 0.81 & 0.79 & 0.79 & 0.78 \\
SynthCLIC & 0.91 & 0.94 & 0.92 & 0.92 \\
\bottomrule
\end{tabular}
\caption{Mean AP (mAP) by dataset and $k$-value for models trained on \synthclic{}. These runs use the augmented factorized-head protocol (Table~\ref{tab:clip_classifier_hyperparameters_detailed}), not the canonical classifier.}
\label{tab:k_ablation_overall}
\end{table}

\subsubsection{Effect of Orthogonality Constraint}
\label{appendix:clip:orthogonality_ablation}

Table~\ref{tab:rev:orthogonality} compares activation- and weight-level
orthogonality regularization for the factorized $k{=}8$ head. We report
\begin{align}
O_W
&=
\left\|
\mathbf{I}
-
\mathbf{W}_1^\top \mathbf{W}_1
\right\|_F,
\label{eq:weight_orthogonality_metric}
\\
O_A
&=
\left\|
\mathbf{I}
-
\bar{\mathbf{A}}^\top \bar{\mathbf{A}}
\right\|_F,
\label{eq:activation_orthogonality_metric}
\end{align}
where $\bar{\mathbf{A}}$ is the column-normalized activation matrix defined
in Appendix~\ref{appendix:clip:orthogonal_head}. The activation regularizer
used during training is $\lambda O_A^2$
(Equation~\ref{eq:activation_orthogonality}). Table~\ref{tab:rev:orthogonality}
reports the corresponding unsquared diagnostics.

Detection performance is unchanged across variants ($0.92$ mAP), but the
regularizers affect different objects. Weight regularization enforces
$O_W=0.00$ while leaving substantial activation non-orthogonality
($O_A=2.64$). Activation regularization directly reduces $O_A$ to $0.43$
and also produces more orthogonal weights ($O_W=1.05$) than using no
auxiliary loss ($O_W=2.25$). This supports activation orthogonality when
the resulting activation dimensions are the objects of analysis.

\begin{table}[!ht]
\centering
\scriptsize
\begin{tabular}{lrrr}
\toprule
Auxiliary loss & mAP & $O_W \downarrow$ & $O_A \downarrow$ \\
\midrule
None       & 0.92 & 2.25 & 6.81 \\
Activation & 0.92 & 1.05 & 0.43 \\
Weight     & 0.92 & 0.00 & 2.64 \\
\bottomrule
\end{tabular}
\caption{Orthogonality ablation on \synthclic{}. Detection performance
is unchanged across variants. Weight regularization enforces orthogonality
of $\mathbf{W}_1$, whereas activation regularization most strongly reduces
the activation-space orthogonality error $O_A$.}
\label{tab:rev:orthogonality}
\end{table}

\subsubsection{CLIP Backbone Ablation}
\label{appendix:clip:backbone_ablation}
Table~\ref{tab:rev:clip_backbones} gives the evidence for the backbone claim of Section~\ref{sec:results:detection_performance}: in-domain mAP on \synthclic{} is similar across the three OpenAI CLIP ViT backbones ($0.89$--$0.92$), with a slight edge for the largest (ViT-L/14-336), and the gap between ViT-B/16 and ViT-B/32 is small and dataset-dependent.

\begin{table}[!ht]
\centering
\scriptsize
\begin{tabular}{lrrr}
\toprule
Test Set (in-domain) & ViT-B/16 & ViT-B/32 & ViT-L/14 \\
\midrule
CNNSpot & 0.93 & 0.87 & 0.96 \\
combined & -- & -- & 0.49 \\
SynthBuster+ & 0.99 & 0.98 & 1.00 \\
SynthCLIC & 0.89 & 0.90 & 0.92 \\
\bottomrule
\end{tabular}
\caption{In-domain mAP across CLIP backbones.}
\label{tab:rev:clip_backbones}
\end{table}

\end{document}